\pdfoutput=1

\documentclass[11pt]{article}

\usepackage[final]{EMNLP2023}

\usepackage{times}
\usepackage{latexsym}
\usepackage{multirow}
\usepackage{booktabs}
\usepackage{algorithm}
\usepackage{algorithmic}
\usepackage{graphicx}
\usepackage{amsmath}
\usepackage{hyperref}
\usepackage{natbib}
\usepackage{amssymb}
\usepackage{graphicx}
\usepackage{colortbl}
\usepackage{xcolor}
\usepackage{tabularx}
\usepackage{adjustbox}

\usepackage[T1]{fontenc}

\usepackage[utf8]{inputenc}

\usepackage{microtype}

\usepackage{inconsolata}

\definecolor{Gray}{gray}{0.95}

\newcommand{\gr}{\rowcolor[gray]{.95}}

\newcommand{\shortercite}[1]{(\citeyear{#1})}

\title{LPZero: Language Model Zero-cost Proxy Search from Zero}

\author{Peijie Dong$^1$, Lujun Li$^2$, Xiang Liu$^1$, Zhenheng Tang$^{3,2}$ \\ \textbf{Xuebo Liu$^4$, Qiang Wang$^4$, and Xiaowen Chu$^{1,2}$} \\
  $^1$HKUST-GZ, $^2$HKUST, $^3$HKBU, $^4$HIT-SZ \\
  \texttt{pdong212@connect.hkust-gz.edu.cn}, 
  \texttt{lilujunai@gmail.com}, \\
  \texttt{xliu886@connect.hkust-gz.edu.cn},
  \texttt{zhtang@comp.hkbu.edu.hk},  \\
  \texttt{\{liuxuebo,qiang.wang\}@hit.edu.cn}, \texttt{xwchu@ust.hk}
}

\begin{document}
\maketitle
\begin{abstract}
In spite of the outstanding performance, Neural Architecture Search (NAS) is criticized for massive computation. Recently, Zero-shot NAS has emerged as a promising approach by exploiting Zero-cost (ZC) proxies, which markedly reduce computational demands. Despite this, existing ZC proxies heavily rely on expert knowledge and incur significant trial-and-error costs. Particularly in NLP tasks, most existing ZC proxies fail to surpass the performance of the naive baseline. To address these challenges, we introduce a novel framework, \textbf{LPZero}, which is the first to automatically design ZC proxies for various tasks, achieving higher ranking consistency than human-designed proxies. Specifically, we model the ZC proxy as a symbolic equation and incorporate a unified proxy search space that encompasses existing ZC proxies, which are composed of a predefined set of mathematical symbols. To heuristically search for the best ZC proxy, LPZero incorporates genetic programming to find the optimal symbolic composition. We propose a \textit{Rule-based Pruning Strategy (RPS),} which preemptively eliminates unpromising proxies, thereby mitigating the risk of proxy degradation. Extensive experiments on FlexiBERT, GPT-2, and LLaMA-7B demonstrate LPZero's superior ranking ability and performance on downstream tasks compared to current approaches.
\end{abstract}

\section{Introduction}



Traditional neural network design~\cite{Krizhevsky2012ImageNetCW}, heavily dependent on expert knowledge and experience~\cite{he2016resnet}, is both time-intensive and prone to trial-and-error. Neural Architecture Search (NAS) emerged to automate and refine this process by identifying optimal architectures from a set of possibilities using various strategies. However, early NAS methods~\cite{zoph2016neural, real2018regularized} require extensive computation, which limits their wide accessibility. For instance,  NASNet~\cite{zoph2016neural} requires 500 GPUs for four days. 


\begin{table}[t]
\centering
\resizebox{.9\linewidth}{!}{
\begin{tabular}{ cc }
\toprule
\textbf{Proxy Name} & \textbf{Formula} \\
\midrule
Activation Distance & $\begin{aligned} \mathcal{S}=\text{log}|K_H| \end{aligned}$\\
\hline
Synaptic Saliency & $\begin{aligned} \mathcal{S}=\frac{\partial \mathcal{L}}{\partial W} \odot W \end{aligned}$\\
\hline
Jacobian Cosine & $\begin{aligned} S = \left[ J_n J_n^t - I \right]^{\frac{1}{20}} \end{aligned}$\\
\hline
Synaptic Diversity & $\begin{aligned} S = \left\| \frac{\partial \mathcal{L}}{\partial W} \right\| \odot \|W\|_{\text{nuc}} \end{aligned}$\\
\hline
Attention Confidence & $\begin{aligned} \mathcal{S} = \max(\text{Att}(h, (x_n))) \end{aligned}$\\
\hline
Softmax Confidence & $\begin{aligned} \mathcal{S} = \max(\text{Sft}(h, (x_n))) \end{aligned}$\\
\hline
Attention Importance & $\begin{aligned} \mathcal{S} = \left|{\partial \text{Att}(I)} \frac{\partial \mathcal{L}(I)}{\partial \text{Att}(I)} \right| \end{aligned}$\\
\hline 
SNIP & $\begin{aligned}\mathcal{S} = \left| \frac{\partial \mathcal{L}}{\partial W} \odot W \right| \end{aligned}$\\
\hline
GraSP & $\begin{aligned} \mathcal{S} = - \left( H \frac{\partial \mathcal{L}}{\partial W} \right) \odot W \end{aligned}$\\
\hline
Fisher & $\begin{aligned}\mathcal{S}=\frac{\partial \mathcal{L}}{\partial A} \times A\end{aligned}$\\
\hline
LogSynflow & $\begin{aligned}\mathcal{S} = W \cdot \left| \log \left| \frac{\partial \mathcal{L}}{\partial W} \right| \right| \end{aligned}$\\
\hline
Synflow & $\begin{aligned} \mathcal{S}=\frac{\partial \mathcal{L}}{\partial \mathcal{W}} \odot W \end{aligned}$\\ 
\hline 
GradNorm & $\begin{aligned} \mathcal{S}=||\frac{\partial \mathcal{L}}{\partial \mathcal{W}}||_F \end{aligned}$\\
\bottomrule
\end{tabular}
}
\caption{Overview of handcrafted Zero-cost proxies for Transformers, notating \( K_H \) as the Kernel Matrix, \( J \) as the Jacobian w.r.t. Mini-Batch Input \( I \), $\text{Att}$ as attention head, $\text{Sft}$ as softmax output, \( A \) as activation, and \( H \) as the Hessian matrix.}\label{table:proxy-revisiting}
\end{table}

To alleviate this issue, recent advancements in Zero-shot NAS~\cite{ZenNAS, li2023zico, mellor2021neural, abdelfattah2021zerocost, ying2019bench, krishnakumar2022bench, DSS} aim to significantly reduce training costs by employing Zero-cost (ZC) proxies, which circumvent the traditional training process and decrease computational demands. Zero-shot NAS predicts the performance of neural network architectures without the need for actual training, using models that are randomly initialized. This approach enables rapid and efficient estimation of architecture performance, eliminating the time and resources typically consumed in training processes.
To evaluate the effectiveness of ZC proxies, Spearman's $\rho$ or Kendall's $\tau$ are utilized to measure the congruence between the performance rankings predicted by ZC proxies and ground truth derived from fully trained models. 
A high-ranking correlation indicates the reliability of ZC proxies in forecasting the potential success of architectures. 

However, existing ZC proxies~\cite{serianni-kalita-2023-training, Javaheripi2022LiteTransformerSearchTN} are heavily dependent on in-depth expert knowledge and a repetitive trial-and-error, which can be both time-intensive and demanding in terms of effort. For instance, Attention Confidence~\cite{serianni-kalita-2023-training} utilizes normalization techniques to refine attention mechanisms for enhanced performance. Meanwhile, pruning-based proxies such as SNIP~\cite{Lee2018SNIPSN}, Fisher~\cite{Turner2019BlockSwapFB_fisher}, GraSP~\cite{Wang2020PickingWT_GraSP}, GradNorm~\cite{abdelfattah2021zerocost} and Synflow~\cite{tanaka2020pruning_synflow} involve complex combination of mathematical operations that critically influence their ranking capabilities. Notably, LogSynflow~\cite{cavagnero2022freerea} implements logarithmic operations to address gradient explosion issues inherent in Synflow. 
Furthermore, we observe that most of the existing proxies cannot surpass the baseline performance, measured by the number of parameters, as presented in Table~\ref{tab:performace_bert} and \ref{tab:performance_gpt2}.
\begin{figure*}[t]
\begin{minipage}[b]{0.45\linewidth}
    \centering
    \includegraphics[width=\linewidth]{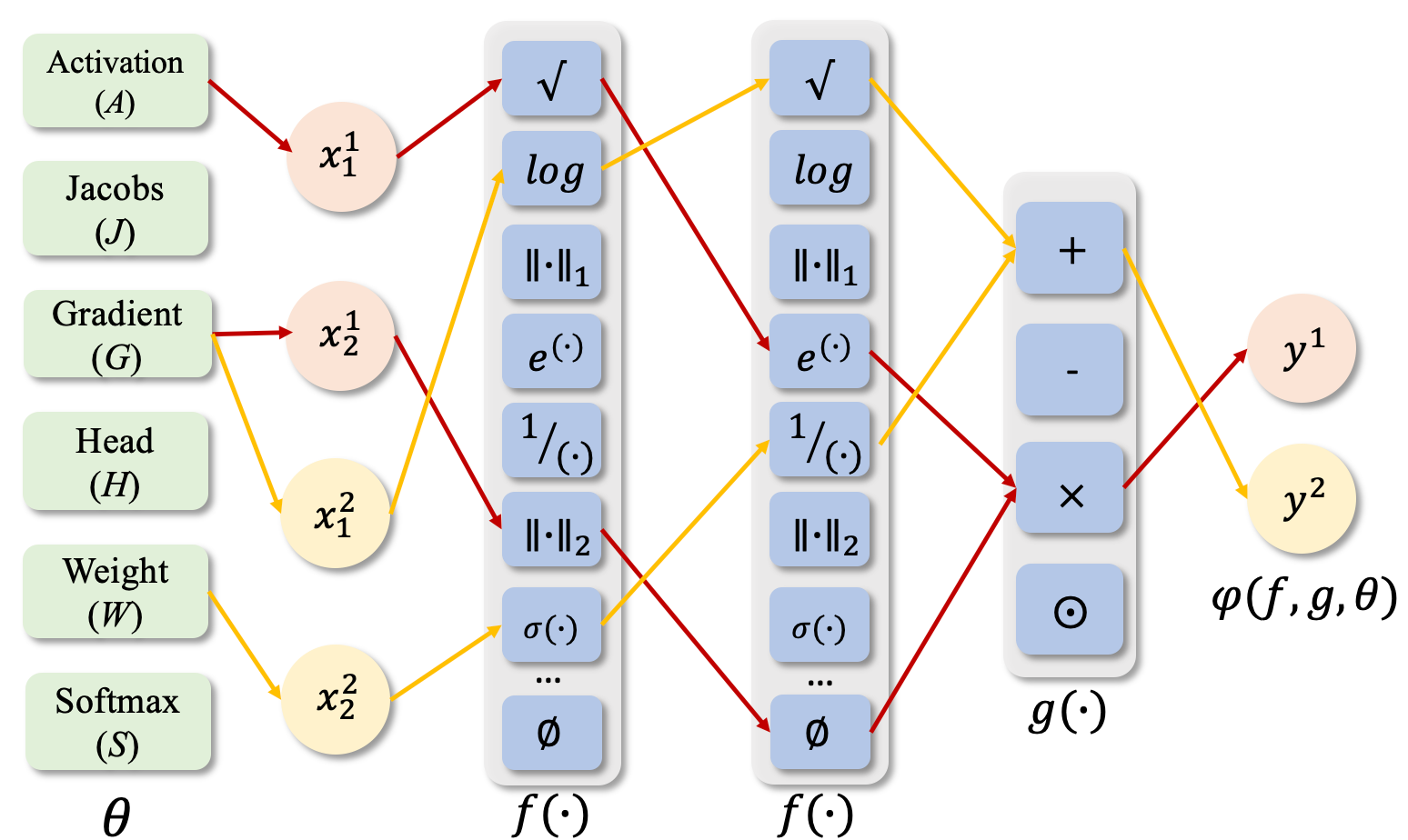}
    \caption{Proxy Search space of LPZero framework.}
    \label{fig:search_space}
\end{minipage}
\begin{minipage}[b]{0.55\linewidth}
    \centering
    \includegraphics[width=\linewidth]{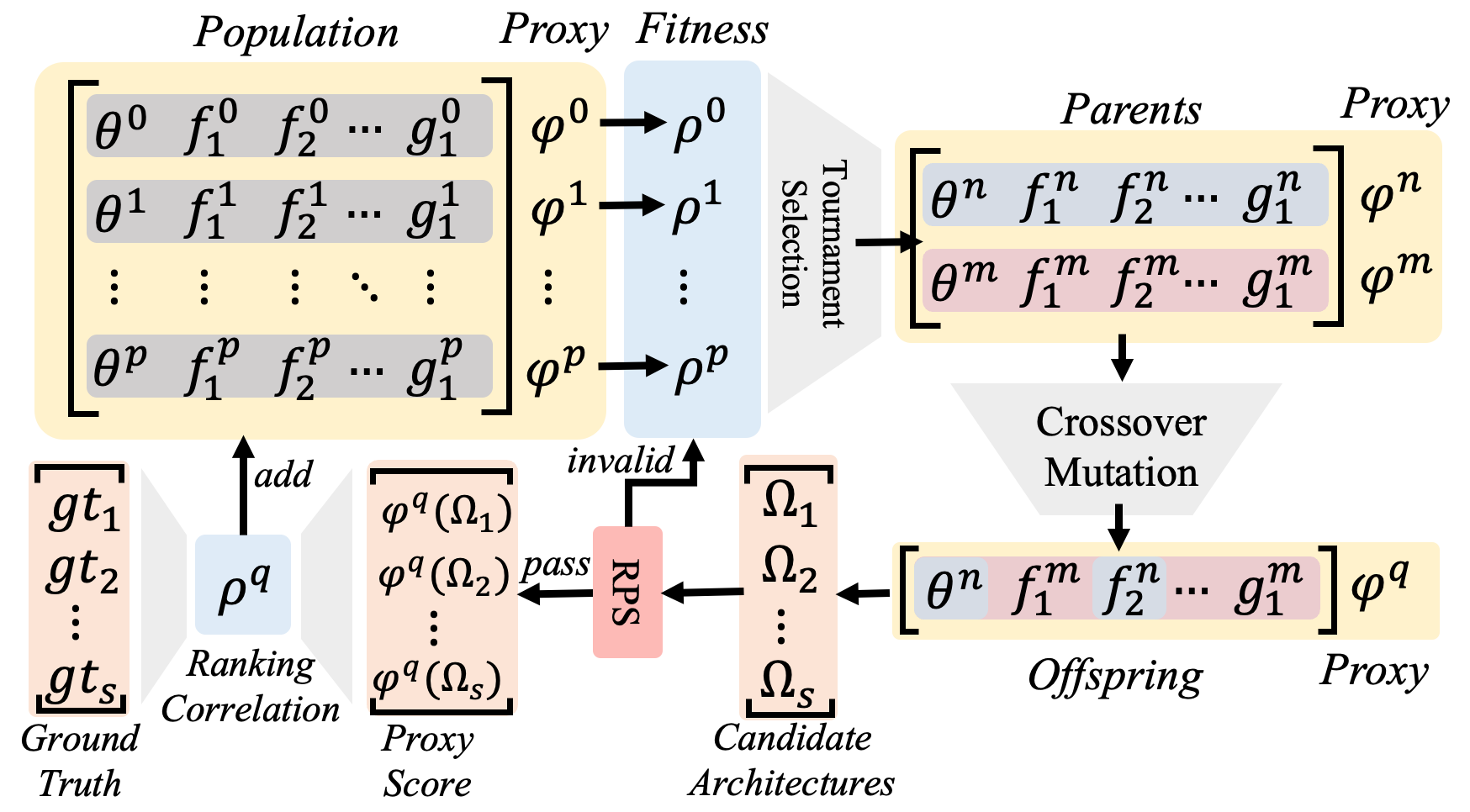}
    \caption{Genetic programming process of LPZero.}
    \label{fig:main_figure}
\end{minipage}
\label{fig:comparison}
\vspace{-0.2in}
\end{figure*}

This limitation raises a fundamental but critical question: 
\textit{\textbf{How to devise new proxies efficiently and automatically for language models?}}

To answer this question, we break it down to two steps: (1) \textbf{Devise a unified proxy search space for existing ZC proxies.} (2) \textbf{Employ genetic programming for discover new proxies. }

For the \textbf{first step}, we revisit the existing ZC proxies, as detailed in Table~\ref{table:proxy-revisiting}, and design a comprehensive proxy search space that encompasses current ZC proxies. Specifically, we formulate the ZC proxies as symbols. Then, these proxies are categorized into six types based on the input type: Activation ($\textit{A}$), Jacobs ($\textit{J}$), Gradients ($\textit{G}$), Head ($\textit{H}$), Weight ($\textit{W}$) and Softmax ($\textit{S}$), illustrated in Figure~\ref{fig:search_space}. Within this unified framework, we select two types of inputs, denoted as $\theta$, from these categories. Each input undergoes transformation through $n$ unary operations $f(\cdot)$, and the results are combined using a binary operation $g(\cdot)$. This process generates a candidate proxy, $\varphi(f, g, \theta)$, for our proxy search space. More details can be found in Appendix~\ref{sec:app:relatedwork}.

For the \textbf{second step}, we propose a \textbf{LPZero} framework, denoting \textbf{L}anguage model \textbf{P}roxy Search from \textbf{Zero}. As illustrated in Figure~\ref{fig:main_figure}, we initially select $p$ candidate proxies to establish the population and assess their ranking consistency within the FlexiBERT benchmark. Through tournament selection, we identify two promising parent proxies ($\varphi^{n,m}$). Subsequently, we perform crossover and mutation operations to generate the offspring proxy $\varphi^q$. To evaluate its ranking consistency Spearman $\rho^q$, we employ this proxy to score each architecture $\Omega_i$ with $\varphi^q(\Omega_i)$ and compare the results with their respective ground truth $\text{gt}_i$ (e.g., average accuracy). Given the sparsity of the proxy search space, we advocate for a \textit{Rule-based Pruning Strategy (RPS)} aimed at eliminating ineffective proxies, thereby enhancing search efficiency. Our main contributions are:
\begin{itemize}\vspace{-0.05in} 
    \item We design a comprehensive and high-quality proxy search space that encompasses most of the existing ZC proxies tailored for language models. To the best of our knowledge, we are the first to present an automatic ZC proxy framework for language models.
    \item We introduce a \textit{Rule-based Pruning Strategy (RPS)} to prevent proxy degradation and improve search efficiency. \vspace{-0.1in}
    \item Experiments on FlexiBERT, GPT-2 and LLaMA substantiate the superiority of the proxies identified by our LPZero, indicating the effectiveness of our proposed approach. 
\end{itemize}

\section{Related Work}\label{sec:related_work}

\noindent\paragraph{Zero-shot NAS} has gained prominence as a computation-efficient alternative to previous NAS methods~\cite{nasnet, darts, ENAS, cai2018proxylessnas}. It can estimate the performance of candidate architectures without extensive training. Existing ZC proxies rely heavily on experts and handcrafted heuristics. For instance,  NWOT~\cite{mellor2021neural} leverages the local Jacobian values across various images to construct an indicator for the model's capability. ZenNAS~\cite{ZenNAS} assesses candidate architectures by employing the gradient norm of input images. Zero-cost NAS~\cite{abdelfattah2021zerocost} introduces pruning-based metrics as ZC proxies, which encompass indicators including GradNorm~\cite{abdelfattah2021zerocost}, SNIP~\cite{Lee2018SNIPSN} and Synflow~\cite{tanaka2020pruning_synflow}, etc. These proxies evaluate the significance of network parameters and aggregate layer-wise values to estimate the overall performance. The above proxies mainly focus on convolution-based networks, recent efforts~\cite{serianni-kalita-2023-training} first apply ZC proxies to transformer-based networks, including RNN and BERT, and propose the FlexiBERT benchmark. LiteTransformerSearch~\cite{Javaheripi2022LiteTransformerSearchTN} proposes to employ the number of decoder parameters as ZC proxies on the GPT-2 benchmark. 

\noindent\paragraph{Automatic Design for ZC Proxies.} Several studies explore how to search for ZC proxies automatically, notably EZNAS~\cite{akhauri2022eznas} and EMQ~\cite{dong2023emq}. EZNAS introduces a proxy search space dedicated to convolution-based networks, achieving commendable performance across various benchmarks~\cite{ying2019bench, dong2019NASBench201}. However, its effectiveness is notably diminished when applied to Transformer-based networks. On the other hand, EMQ~\cite{dong2023emq} develops a specialized proxy search space tailored for mixed-precision quantization proxies for Convolution-based networks. Our LPZero framework can be applied to Transformer-based architectures, particularly language models, and shows superior and more promising performance.

\noindent\paragraph{NAS for LLMs.} Large Language Models (LLMs), such as LLaMA~\cite{sarah2024llama}, are becoming increasingly large, with model sizes ranging from 7B to 70B~\cite{zhang2023dissectingruntimeperformancetraining}. This rapid growth in model scale poses challenges for directly applying supernet-based NAS methods~\cite{cai2019once,yu2020bignas} to LLMs. To address this issue, recent works, including LoNAS~\cite{munoz-etal-2024-lonas-elastic} and LLaMA-NAS~\cite{sarah2024llama}, leverage elastic Low-Rank Adaptation (LoRA)~\cite{hu2022lora} to transform a pre-trained LLM into a supernet. This approach enables the practical application of NAS techniques to LLMs by reducing the search space and computational requirements. In this paper, we further enhance the efficiency of the sub-network search process by employing LPZero as a cost-effective performance estimator.


\section{Methodology}

\subsection{Proxy Search Space}

The search spaces of most AutoML approaches~\cite{real2020automlzero, darts} are specifically designed for particular purposes and not suitable for proxy search. 
Previous auto loss search methods~\cite{li2020auto, Li2021AutoLossZeroSL, Gu2022AutoLossGMSSG} take the output of network $y$ and ground truth $\hat{y}$ as input (scalar), which is relatively easy to handle. However, for the ZC proxies search problem, we involve more operations that take scalar, vector, and matrix as input, which might deduce the shape mismatching problem. The complete operations are presented in Table~\ref{tab:operation_vocabulary} of Appendix~\ref{sec:app:proxy_ss}.

LPZero aims to identify the most suitable ZC proxy to accurately assess network performance. The primary objective is to optimize Spearman's rank correlation coefficient ($\rho$), which measures the ranking consistency of each ZC proxy. Thus, our training-free approach is formulated as follows:
\begin{equation}
\varphi^* = \underset{\varphi \in \mathcal{S}}{\text{argmax}} (\rho(\varphi)), \
\varphi = \varphi(f, g, \theta).
\end{equation}
where $\varphi$ represents the candidate ZC proxies within the proxy search space $\mathcal{S}$. Each proxy $\varphi$ is defined as a function of unary and binary operations ($f$ and $g$) applied to input parameters $\theta$.

\begin{figure}[t]
    \centering
    \includegraphics[width=1\linewidth]{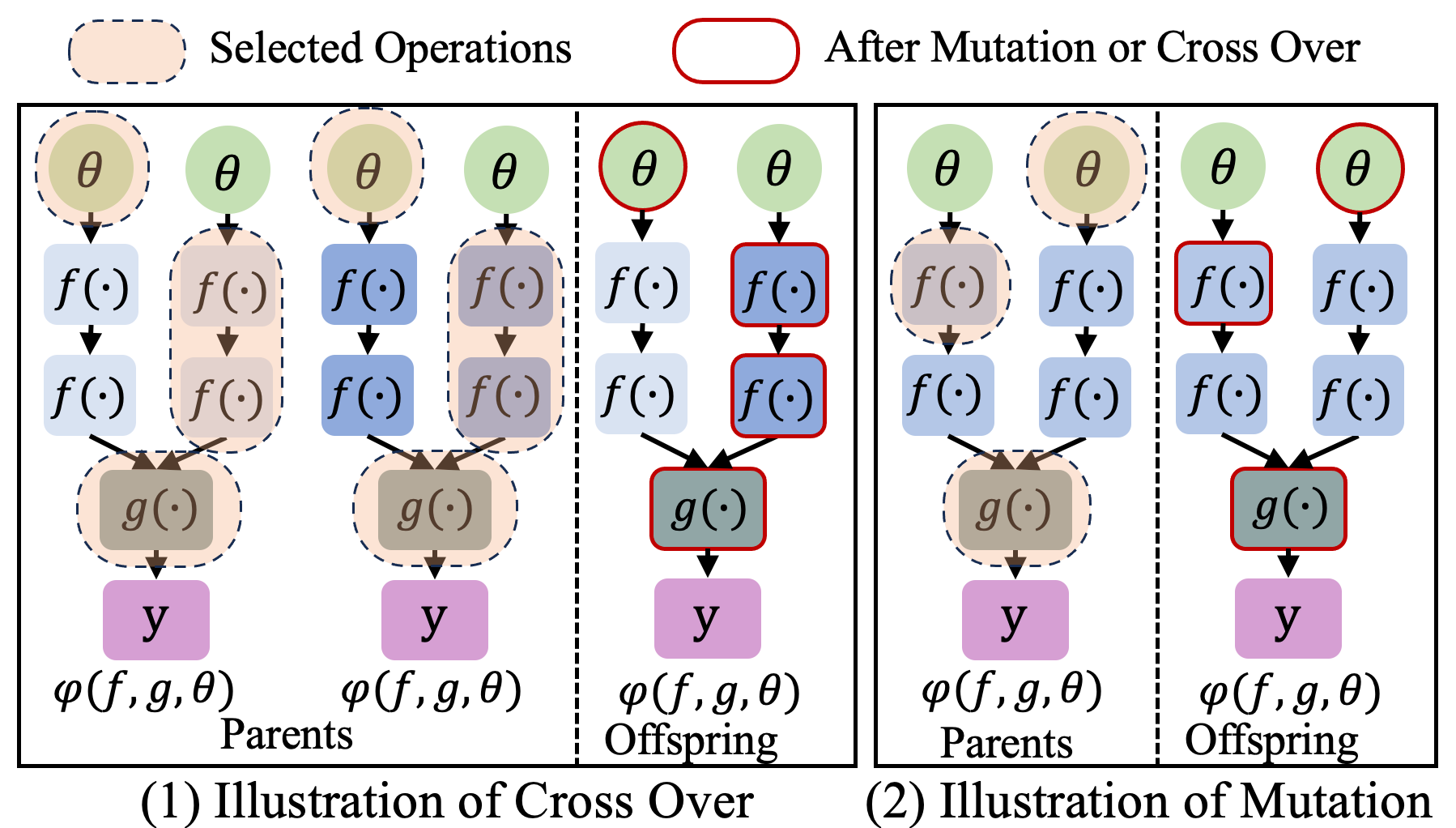}
    \caption{Illustration of Crossover and Mutation.}
    \label{fig:crossover_mutation}
\end{figure}

\begin{algorithm}[t]
\small 
\caption{LPZero Algorithm}
\label{alg:gene_prog}
\begin{algorithmic}[1]
\STATE \textbf{Input:} Initial population size $p$, number of generations $G$, crossover rate $C_r$, mutation rate $M_r$
\STATE \textbf{Output:} ZC proxy with highest Spearman

\STATE Initialize population with $p$ random ZC proxies
\FOR{$g = 1$ to $G$}
    \STATE Evaluate fitness of each proxy in the population
    \STATE Pick top $\mathcal{R}$ ratio as pool $\mathcal{Q}$ 
    \STATE Select parents $\varphi^{n,m}$ randomly from $\mathcal{Q}$ 
    \STATE CrossOver $\varphi^q=\text{CrossOver}(\varphi^{n}, \varphi^{m})$ with probability $C_r$.
    \STATE Mutation $\varphi^q=\text{Mutate}(\varphi^q)$ with probability $M_r$
    \IF{RPS($\varphi^q$) is valid}
        \STATE Add offspring to population
    \ELSE
        \STATE Jump to Line 8 and regenerate offspring $\varphi^{q}$
    \ENDIF
    \STATE Evaluate fitness of new offspring $\varphi^q$
    \STATE Keep the top-$p$ proxies for the next generation
\ENDFOR
\STATE \textbf{return} the proxy with the highest Spearman
\end{algorithmic}
\end{algorithm}

\noindent\textbf{Zero-cost Proxy Representation.} The ZC proxy $\varphi$ is represented as a Symbolic Expression (SE). As illustrated in Figure~\ref{fig:search_space}, the algorithmic expression can be represented by the combination of unary operations $f(\cdot)$ and binary operations $g(\cdot)$. Therefore, SE can be represented as $\varphi(f, g, \theta)$, where inputs $x_1$ and $x_2$ is chosen from six candidates $\theta$, including Activation ($\textit{A}$), Jacobs ($\textit{J}$), Gradients ($\textit{G}$), Head ($\textit{H}$), Weight ($\textit{W}$) and Softmax ($\textit{S}$). 

\noindent\textbf{Primitive Operations.} Table~\ref{tab:operation_vocabulary} summarizes the primitive operation set $\mathcal{K}$ used in the proxy search space. This set comprises 20 unary operations and four binary operations, facilitating information exchange across dimensions. These operations are non-parametric, meaning they do not have adjustable parameters, making them highly efficient and effective in various computational tasks. Unary operations act on a single input, while binary operations operate on pairs of inputs. Notably, $f_{19}$ and $f_{20}$ are unique unary operations; $f_{20}$ signifies a pass-through where the input is returned without any modification, and $f_{20}$ represents a pruning operation that results in the removal of the branch, effectively returning nothing. By incorporating this diverse set of operations, our proxy search space can explore a wide range of function transformations, enabling the discovery of novel architectures and enhancing the flexibility of our approach.

\noindent\textbf{Analysis for the Proxy Search Space.} In Figure~\ref{fig:search_space}, we illustrate the proxy search space by showcasing two proxies depicted in red and yellow lines, demonstrating the variability and richness of architectural configurations. With a total of 20 unary operations and 4 binary operations available, the proxy search space is expansive, yielding a combinatorial space of $C_6^2 \times 20^2 \times 4 = 24,000$ potential ZC proxies. This vast space enables the exploration of a wide spectrum of architectural designs, allowing for the discovery of innovative solutions tailored for the specific requirements of NLP tasks. 

\subsection{LPZero Framework} 

Inspired by the AutoML~\cite{He2021AutoMLAS, li2019lfs}, genetic programming is employed as the core mechanism for our search algorithm. It leverages the principles of natural selection and genetic evolution to optimize models and hyperparameters. Figure~\ref{fig:main_figure} illustrates the search pipeline of our LPZero framework. At initialization, we uniformly sample $p$ ZC proxies from the proxy search space to get the initial population. Then, we measure the ranking correlation on the architecture search space to measure the predictability of each proxy. Then, for each iteration, we conduct tournament selection to pick $\mathcal{R}$ ratios from a population ($\mathcal{R}=10\%$ by default) as promising candidates, and then randomly sample two of them as parents $\varphi^{n,m}$. Then, the parents are utilized to perform crossover and mutation with a probability of $C_r$ and $M_r$ respectively to get the offspring. To verify the effectiveness of offspring, we sample $S$ candidate architectures from the architecture search space and compute the ranking correlation of ground truth and proxy score. As the proxy search space is very sparse with a large number of unpromising or even invalid ZC proxies, we propose early-stopping to filter out the candidates. 

\noindent\textbf{Crossover and Mutation.} Each symbolic expression consists of two branches and one aggregation node. These branches represent the individual components or operations within the proxy architecture, while the aggregation node combines the outputs of these branches to form the final proxy score. As shown in Figure~\ref{fig:crossover_mutation}, we present the illustration of CrossOver and Mutation. During the crossover operation, two parents are selected, and genetic information is exchanged between them to generate offspring. This process involves swapping segments of the parents to create new combinations of operations and architectures. Conversely, the mutation operation introduces random alterations to the genetic makeup of a single SE, potentially introducing novel architectures into the population.

\noindent\textbf{Rule-based Pruning Strategy.} The Rule-based Pruning Strategy in the LPZero framework serves a crucial role in managing the computational challenges posed by the expansive and sparsely populated proxy search space. It works to promptly identify and discard unpromising or invalid ZC proxies, thereby conserving computational resources and expediting the search for optimal solutions. By utilizing predefined criteria as presented in Appendix~\ref{app:predefined} this strategy evaluates the viability of candidate proxies. Those failing to meet the specified criteria are removed from the population, reducing the proxy search space and focusing computational efforts on promising candidates. Overall, this strategic filtering process enhances the efficiency and effectiveness of the LPZero framework, facilitating swifter progress toward the discovery of high-quality proxy architectures.

\noindent\textbf{Searched Zero-cost Proxy.} Based on the LPZero framework, we present the searched ZC proxy tailored for the different tasks, including GPT-2, FlexiBERT, and LLaMA benchmark, characterized by a unique combination of structural and operational elements. The architecture of this proxy is delineated as follows: the input structure comprises heads and activation functions, and the tree structure utilizes operations such as element-wise reversion, element-wise power, Frobenius norm, log softmax, etc. For more operations, refer to Table~\ref{tab:operation_vocabulary}. For these three tasks, we present the searched ZC proxies in Appendix~\ref{app:searched_proxy}.

\section{Experiments}

In this section, we first detail the experimental setup and implementation details of LPZero. Then, we present the ranking correlation evaluation on FlexiBERT and GPT-2 benchmark. Subsequently, we assess LPZero's performance by examining the ranking correlation in the FlexiBERT and GPT-2 benchmarks. After that, we report the performance on commonsense tasks for LLaMA-7B model. Finally, we conduct an ablation study to evaluate the impact of our genetic programming framework, the \textit{Rule-based Pruning Strategy (RPS)}, and other components including the number of unary operations and the initial population size.

\begin{figure*}[t]
    \centering
    \includegraphics[width=.95\linewidth]{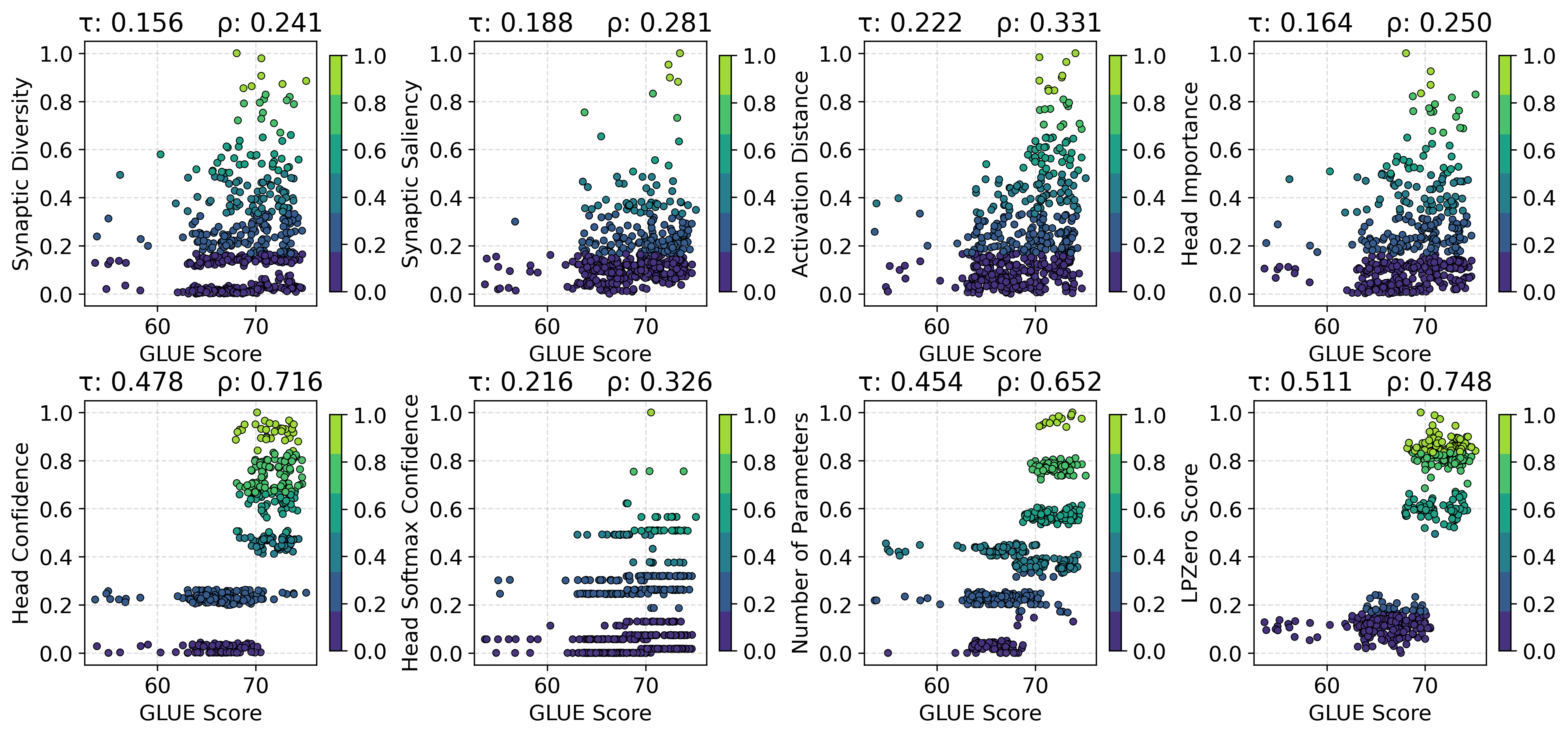}
    \caption{Spearman's $\rho$ and Kendall's $\tau$ Correlation of training-free proxies with GLUE Score across 500 architectures randomly sampled from FlexiBERT benchmark.}
    \label{fig:correlation-training-free}
\end{figure*}

\subsection{Implementation Details} 

\noindent\textbf{Datasets.} FlexiBERT~\cite{serianni-kalita-2023-training} is built on the GLUE benchmark~\cite{Wang2018GLUEAM}. We adopt the average performance of the tasks as ground truth to measure the ranking consistency. We employ OpenWebText~\cite{Gokaslan2019OpenWeb} to search for ZC proxies on the FlexiBERT benchmark. For the GPT-2 benchmark, we conduct experiments on the WikiText-103 dataset~\cite{merity2016pointer}. For LLaMA, we conduct experiments on eleven commonsense reasoning datasets: BoolQ~\cite{Clark2019BoolQET}, PIQA~\cite{bisk2020piqa}, HellaSwag~\cite{Zellers2019HellaSwagCA}, WinoGrande~\cite{Sakaguchi2019WinoGrande}, ARC~\cite{Clark2018ThinkYH} and OBQA~\cite{Mihaylov2018CanAS}.

\noindent\textbf{Criteria.} The effectiveness of ZC proxies is measured by Kendall's $\tau$ and Spearman's $\rho$, with values from -1 to 1, where higher values indicate that the proxies accurately predict the rankings of neural architectures compared to fully trained models. For commonsense reasoning tasks, we employ accuracy as criterion. 

\begin{table}[t]
\centering
\resizebox{.9\linewidth}{!}{
\begin{tabular}{lcc}
\toprule
\textbf{Proxy Name} & \textbf{$\tau$} & \textbf{$\rho$} \\ \midrule
Synaptic Diversity~\shortercite{DSS} & 0.021 & 0.175 \\
Head Importance~\shortercite{serianni-kalita-2023-training} & 0.050 & 0.171 \\ 
Activation Distance~\shortercite{mellor2021neural} & 0.081 & 0.123 \\
Jacobian Cosine~\shortercite{celotti2020improving} & 0.116 & 0.149 \\
SNIP~\shortercite{Lee2018SNIPSN} & 0.119 & 0.173 \\
GraSP~\shortercite{Wang2020PickingWT_GraSP} & 0.122 & 0.179 \\
GradNorm~\shortercite{abdelfattah2021zerocost} & 0.133 & 0.197 \\ 
Fisher~\shortercite{Turner2019BlockSwapFB_fisher} & 0.139 & 0.209 \\ 
Synaptic Saliency~\shortercite{tanaka2020pruning_synflow} & 0.157 & 0.266 \\
Synflow~\shortercite{tanaka2020pruning_synflow} & 0.322 & 0.471 \\
LogSynflow~\shortercite{cavagnero2022freerea} & 0.334 & 0.491 \\
No.Params.~\shortercite{abdelfattah2021zerocost} & 0.454 & 0.590 \\
Attention Confidence~\shortercite{serianni-kalita-2023-training} & 0.475 & 0.666 \\
EZNAS~\shortercite{akhauri2022eznas} & 0.483 & 0.698 \\ 
\gr LPZero (Ours) & \textbf{0.511} & \textbf{0.748} \\ 
\bottomrule
\end{tabular}}
\caption{Ranking correlation of Zero-cost proxies on the FlexiBERT benchmark over 500 architectures with Kendall's $\tau$ and Spearman's $\rho$.}\label{tab:performace_bert}
\end{table}

\begin{table}[t]
\centering
\resizebox{0.88\linewidth}{!}{
\begin{tabular}{lcc}
\toprule
\textbf{Proxy Name} & \textbf{$\tau$} & \textbf{$\rho$} \\ \midrule
Jacobian Cosine~\shortercite{celotti2020improving} & 0.227 & 0.362 \\
EZNAS~\shortercite{akhauri2022eznas} & 0.489 & 0.704 \\
No.Params~\shortercite{abdelfattah2021zerocost} & 0.582 & 0.737 \\
Synflow~\shortercite{tanaka2020pruning_synflow} & 0.632 & 0.730 \\
Activation Distance~\shortercite{mellor2021neural} & 0.644 & 0.818 \\
Attention Confidence~\shortercite{serianni-kalita-2023-training} & 0.676 & 0.850 \\
Fisher~\shortercite{Turner2019BlockSwapFB_fisher} & 0.691 & 0.872 \\
GraSP~\shortercite{Wang2020PickingWT_GraSP} & 0.765 & 0.922 \\
GradNorm~\shortercite{abdelfattah2021zerocost} & 0.834 & 0.958 \\
LogSynflow~\shortercite{cavagnero2022freerea} & 0.836 & 0.962 \\
Synaptic Diversity~\shortercite{DSS} & 0.841 & 0.957 \\
Decoder.Params~\shortercite{Javaheripi2022LiteTransformerSearchTN} & 0.847 & 0.967 \\
Synaptic Saliency~\shortercite{tanaka2020pruning_synflow} & 0.855 & 0.970 \\
SNIP~\shortercite{Lee2018SNIPSN} & 0.858 & 0.970 \\
Head Importance~\shortercite{serianni-kalita-2023-training} & 0.861 & 0.971 \\
\gr LPZero (Ours) & \textbf{0.886} & \textbf{0.980} \\
\bottomrule
\end{tabular}
}
\caption{Ranking correlation of Zero-cost proxies on the GPT-2 benchmark over 200 architectures with Kendall's $\tau$ and Spearman's $\rho$.}\label{tab:performance_gpt2}
\end{table}

\noindent\textbf{Benchmarks.} We employ two language benchmarks to measure the ranking consistency, including FlexiBERT and GPT-2 Benchmark. FlexiBERT Benchmark~\cite{serianni-kalita-2023-training} is a challenging benchmark that encompasses over $10^7$ architectures (Refer to Appendix~\ref{sec:app:flexibert}). We adopt the GPT-2 Benchmark~\cite{Javaheripi2022LiteTransformerSearchTN} on WikiText-103, which provides $10^{54}$ architectures (Refer to Appendix~\ref{sec:app:gpt-2}). For LLaMA search space, we follow the settings in LoNAS and list the details in Appendix~\ref{sec:app:llama_ss}

\noindent\textbf{Genetic Programming Settings.} The configuration of our genetic programming algorithm is as follows: The total number of generations, denoted as $G$, is established at 1,000, with the initial population size, $p$, set to 80 individuals. The probabilities for crossover and mutation operations are both set at $C_r=0.5$ and $M_r=0.5$, respectively. The selection pressure, represented by the ratio $\mathcal{R}$, is fixed at 10\%. A consistent seed of 42 is utilized to ensure reproducibility across experiments. All experiments are conducted on A6000 and H800. During genetic programming, we only require a mini-batch of input (batch size of 128, 16, 32 for BERT, GPT-2 and LLaMA) to calculate the input statistics.

Following EZNAS~\cite{akhauri2022eznas}, we assess the ranking consistency by sampling 50 architectures. Upon finalizing the search proxy, we proceed to evaluate its performance by applying it to two distinct datasets: FlexiBERT, comprising 500 architectures, and GPT-2, encompassing 200 architectures. The whole search process requires 10 GPU hours. 

\noindent\textbf{Training and Evaluation.} We leverage the open-source code by \citet{serianni-kalita-2023-training} and \citet{abdelfattah2021zerocost} to implement the FlexiBERT and various proxies as shown in Table~\ref{table:proxy-revisiting}. We further use the source code in \citet{Javaheripi2022LiteTransformerSearchTN} to implement the GPT-2 benchmark and we collect the benchmark data from their open-sourced repository.
To assess ranking consistency, we sample 500 architectures from the FlexiBERT benchmark, with findings presented in Table~\ref{tab:performace_bert}. Similarly, for the GPT-2 benchmark, we sample 200 architectures to evaluate their ranking consistency, as detailed in Table~\ref{tab:performance_gpt2}.

\begin{figure}[t]
    \centering
    \includegraphics[width=.9\linewidth]{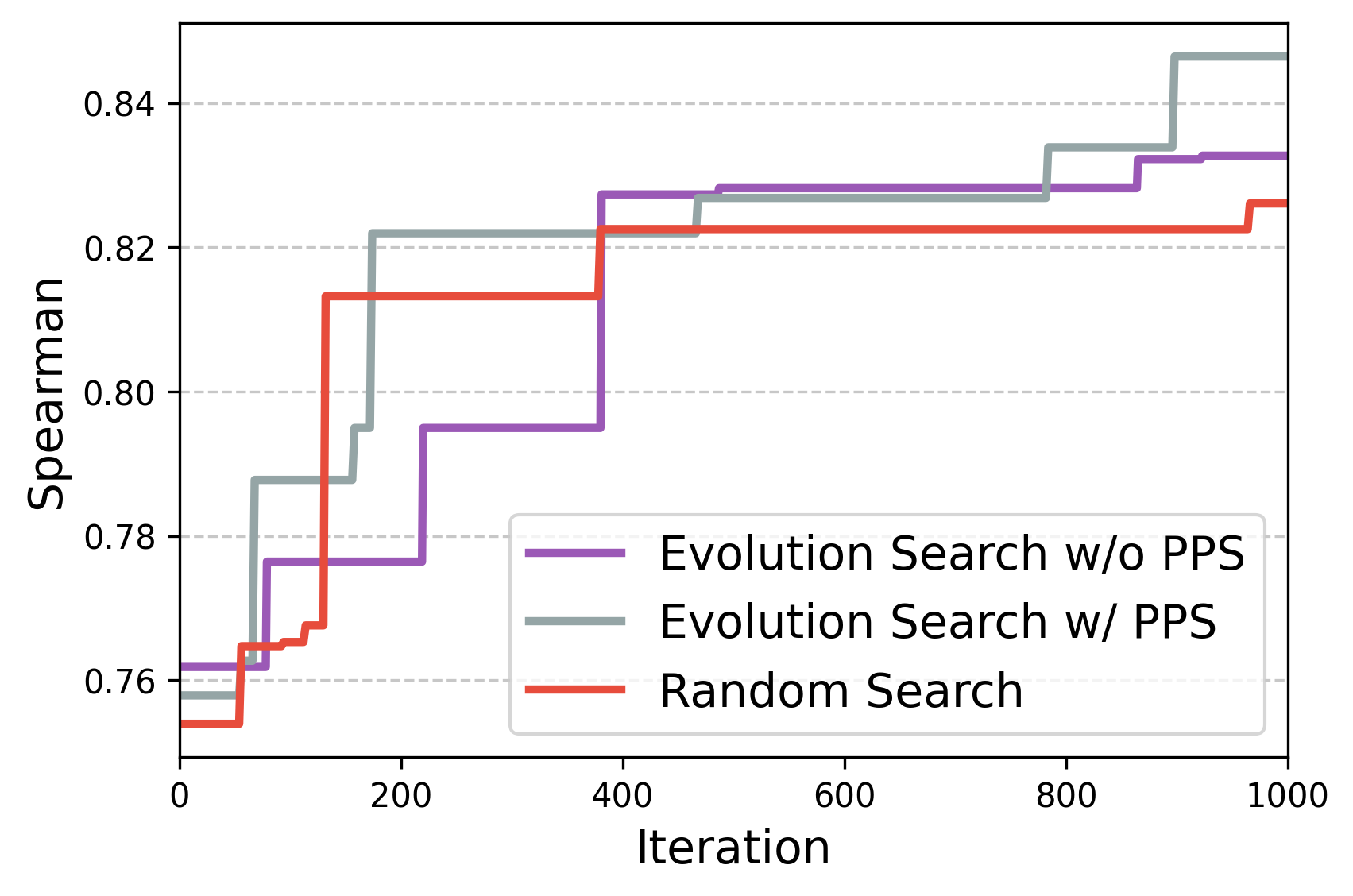}
    \caption{Performance comparison of evolution search with and without the Rule-based Pruning Strategy (RPS) and random search across iterations.}
    \label{fig:evolution_search_process}
    \vspace{-0.1in}
\end{figure}

\begin{figure}[t]
    \centering
    \includegraphics[width=.9\linewidth]{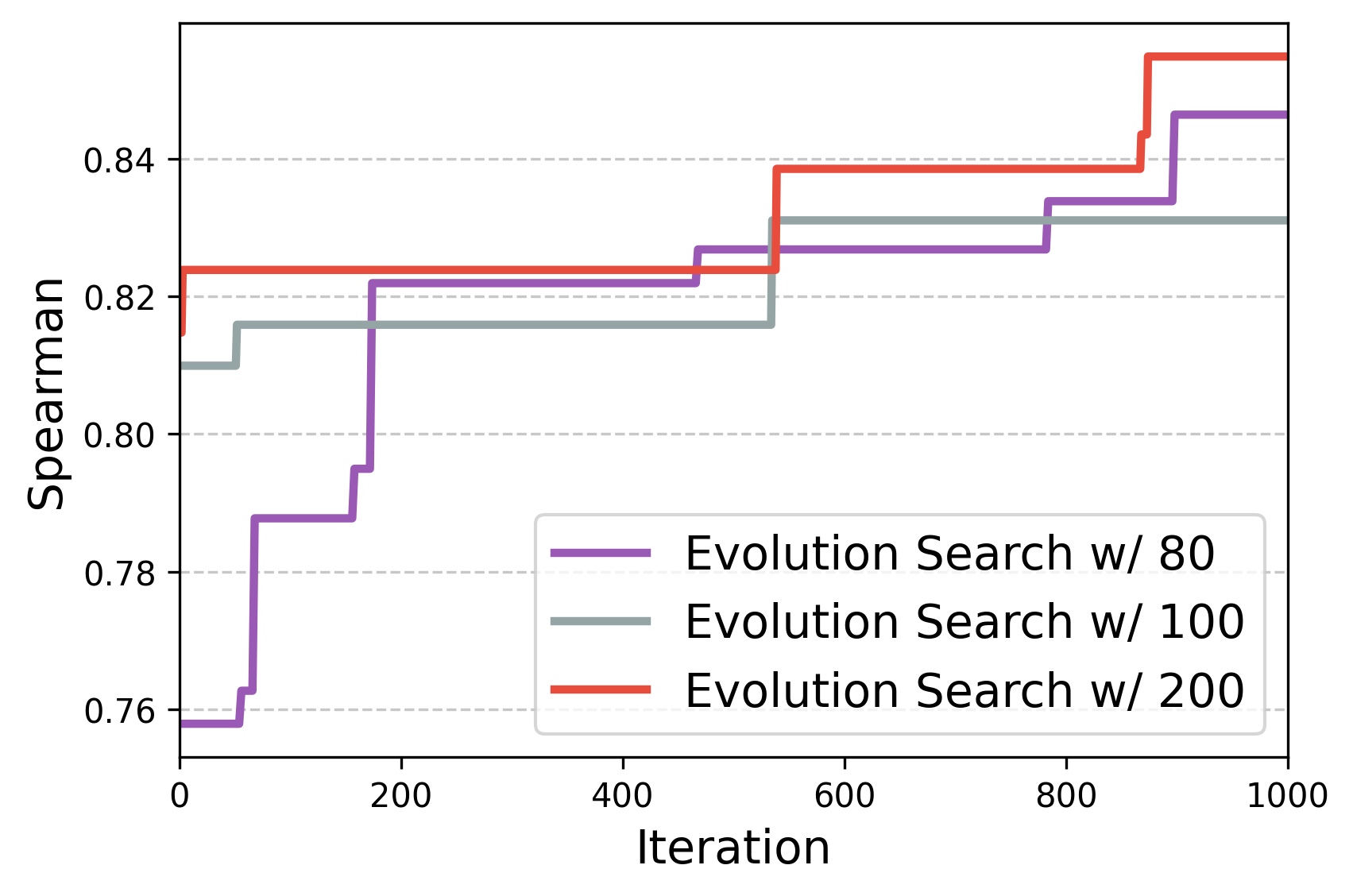}
    \caption{Performance comparison of different size of population.}
    \label{fig:evolution_population}
    \vspace{-0.1in}
\end{figure}


\subsection{Ranking Evaluation} 

\paragraph{Performance on FlexiBERT}

As illustrated in Table~\ref{tab:performace_bert}, we benchmark Kendall's $\tau$ and Spearman's $\rho$ of 14 ZC proxies over 500 architectures from the FlexiBERT benchmark. The baseline (number of parameters) serves as a competitive counterpart and most of the proxies fail to surpass the baseline.
Our LPZero model demonstrates superior ranking consistency, as evidenced by the values of $\tau=0.51$ and $\rho=0.75$ for the respective coefficients. Furthermore, we elucidate the correlation between GLUE scores and ZC proxies through Figure~\ref{fig:correlation-training-free}, which contrasts LPZero with the existing ZC proxies~\cite{serianni-kalita-2023-training} in their study on training-free evaluation methods. This comparison clearly illustrates that our methodology exhibits the highest-ranking consistency among the evaluated frameworks. For additional experiments on FlexiBERT, refer to Appendix~\ref{app:flexibert_exp}.

\begin{table*}[t]
\centering
\resizebox{\linewidth}{!}{
\begin{tabular}{lccccccccc}
\toprule 
\textbf{Method} & \textbf{Params.} & \textbf{BoolQ} & \textbf{PIQA} & \textbf{HellaSwag} & \textbf{WinoGrande} & \textbf{Arc-e} & \textbf{Arc-c} & \textbf{OBQA} & \textbf{Avg} \\ 
\midrule 
LLaMA~\shortercite{sarah2024llama} & 6.7B & 73.18 & 78.35 & 72.99 & 67.01 & 67.45 & 41.38 & 42.40 & 63.25 \\ \hline 
SliceGPT~\shortercite{slicegpt_iclr24} & 5.0B & - & 66.87 & 63.38 & 54.16 & 58.46 & 34.56 & - & 55.49 \\
LLM-Pruner~\shortercite{Ma2023LLMPrunerOT} & 5.4B & 59.39 & 75.57 & 65.34 & 61.33 & 59.18 & 37.12 & 39.80 & 56.82 \\
FLAP~\shortercite{an2023fluctuationbased} & 5.5B & 69.63 & \textbf{76.82} & \textbf{71.20} & \textbf{68.35} & 69.91 & 39.25 & 39.40 & 62.08 \\
SLEB~\shortercite{Song2024SLEBSL} & 5.5B & - & 73.07 & 58.96 & 62.47 & 56.48 & 33.02 & - & 56.80 \\
Shortened LLaMA~\shortercite{kim2024shortened} & 5.5B & \textbf{72.70} & 75.70 & 70.40 & 63.60 & 69.50 & 40.10 & 41.20 & 61.89 \\
\gr LPZero (Ours) & 5.7B & 62.17 & 74.81 & 52.65 & 65.43 & \textbf{74.62} & \textbf{57.42} & \textbf{62.17} & \textbf{64.18} \\ 
\bottomrule
\end{tabular}}
\caption{Performance comparison of different structured pruning-based methods on downstream tasks for LLaMA-7B. All of methods are conducted based on LLaMA, and we report the performance of LLaMA as baseline. ``-'' denotes the data is not available in papers. }
\label{tab:method_comparison}
\end{table*}

\begin{table*}[htbp]
\centering
\label{tab:your_label}
\resizebox{\linewidth}{!}{
\begin{tabular}{lccccccccccc}
\toprule 
\textbf{Method} & \textbf{Params.} & \textbf{TFLOPs} & \textbf{BoolQ} & \textbf{PIQA} & \textbf{HellaSwag} & \textbf{WinoG} & \textbf{Arc-e} & \textbf{Arc-c} & \textbf{OBQA} & \textbf{Avg} & \textbf{Search Time}\\
\midrule
LoNAS-SuperNet & 6.7B & 1.72 & 65.8 & 77.3 & 57.0 & 67.6 & 79.0 & 61.9 & 77.4 & 69.4 & - \\ \hline 
LoNAS-SubNet & 5.6B & 1.44 & \textbf{62.9} & 73.0 & 51.4 & 63.9 & 72.3 & \textbf{58.5} & \textbf{71.0} & \textbf{64.7} & 2.5 GPU hours \\
\gr LoNAS-LPZero (Ours) & 5.7B & 1.46 & 62.2 & \textbf{74.8} & \textbf{52.7} & \textbf{65.4} & \textbf{74.6} & 57.4 & 62.2 & 64.2 & 0.5 GPU hours \\
\bottomrule 
\end{tabular}}
\caption{Comparison of search efficiency with LoNAS for the LLaMA-7B model. The search time reported does not include the evaluation time. LoNAS-SuperNet represents the maximum subnet within the LLaMA model, serving as the basis for LoNAS-SubNet and LoNAS-LPZero experiments.}
\label{tab:search_efficiency_comparison}
\vspace{-0.1em}
\end{table*}

\paragraph{Performance on GPT-2}

As illustrated in Table~\ref{tab:performance_gpt2}, we benchmark Kendall's $\tau$ and Spearman's $\rho$ of 15 ZC proxies over 200 randomly sampled architectures from the GPT-2 benchmark. The additional proxy ~\cite{Javaheripi2022LiteTransformerSearchTN} is ``Decoder.Params'', which represent the parameter of the decoder in GPT-2 models.  Our LPZero achieves the SOTA performance among all ZC proxies, achieving $\tau=0.87$ and $\rho=0.98$. Compared with the FlexiBERT benchmark, the ranking consistency is much higher than the GPT-2 benchmark. 


\subsection{Experiments on LLaMA}

Due to the substantial computation burden of LLMs, training a LLaMA model from scratch is impractical. Inspired by LoNAS~\cite{munoz-etal-2024-lonas-elastic}(under MIT license), we utilize low-cost LoRA as adapters to convert a pre-trained LLM into a weight-sharing super-network. After getting the super-network, LoNAS explores the sub-networks by maximizing the heuristics. However, it is also expensive to evaluate the performance of subnets on downstream tasks. For instance, evaluating a sub-network performance on all downstream tasks in Table~\ref{tab:method_comparison} requires approximately one hour. Given a search space of $2^{31}\times 5^{31}$ proxies, it is infeasible to evaluate all possible sub-networks. Our LPZero method significantly alleviates this issue. 
It serves as a cost-effective estimator for the performance of downstream tasks, requiring only a single forward pass. As presented in Table~\ref{tab:method_comparison}, the results of sub-networks identified using the LPZero proxy surpass those of other counterparts to some extent. In this experiment, we incorporate the LPZero framework into LoNAS for efficient search.

We primarily compare structured pruning methods as baseline, including LoNAS~\cite{munoz-etal-2024-lonas-elastic}, LLM-Pruner~\cite{Ma2023LLMPrunerOT}, SliceGPT~\cite{slicegpt_iclr24}, Wanda~\cite{Sun2023ASA_wanda}, FLAP~\cite{an2023fluctuationbased}, SLEB~\cite{Song2024SLEBSL}, and Shortened LLaMA~\cite{kim2024shortened}. Structured pruning methods can be regarded as an approach to identifying subnetworks within a pre-trained neural network. Consequently, we have chosen these methods for comparison. Our LPZero method exhibits satisfactory performance relative to these counterparts.

Additionally, we present a comparison with the SuperNet-based NAS method LoNAS to show the search efficiency of LPZero. As shown in Table~\ref{tab:search_efficiency_comparison},
LoNAS requires 2.5 GPU hours to search for a subnet, achieving an average score of 64.7. In contrast, In contrast, LPZero requires only 0.5 GPU hours, achieving a similar average score of 64.2. This indicates that LPZero can significantly reduce the evaluation time, which is particularly beneficial when the evaluation process is time-consuming. For the efficiency of proxies, refer to Appendix~\ref{app:efficiency_zc_proxies}.

\subsection{Ablation Study}


\begin{table}[t]
\centering
\resizebox{1\linewidth}{!}{
\begin{tabular}{lcccc}
\toprule
\textit{\#Unary}   & 2       & 3       & 4       & 5       \\ \midrule
Spearman's $\rho$ & \textbf{86.48\%} & 77.47\% & 75.15\% & 78.12\% \\
Winning Rate & \textbf{25.61\%} & 7.96\%  & 8.69\%  & 6.25\% \\ \bottomrule 
\end{tabular}}
\caption{Influence of the Number of Unary Operations on Spearman's $\rho$ and Winning Rate.}\label{tab:ablation_unary_number}
\vspace{-0.1in}
\end{table}

\noindent\textbf{Effectiveness of Genetic Programming.} 
As depicted in Figure~\ref{fig:evolution_search_process}, we limit the number of iterations to 1,000, maintaining an initial population size of 80 throughout the process. The findings reveal that the Evolutionary Algorithm substantially surpasses the performance of Random Search. This indicates that the evolutionary algorithm can heuristically enhance the speed of the search process, thereby significantly improving search efficiency.

\noindent\textbf{Effectiveness of Rule-based Pruning Strategy~(RPS).} As shown in Figure~\ref{fig:evolution_search_process}, we present the performance of the RPS. Our findings indicate that for iterations fewer than 400, RPS not only achieves higher Spearman's $\rho$ but also significantly outperforms evolutionary search methodologies not incorporating RPS, highlighting its critical role in enhancing search efficiency.

\noindent\textbf{Initial Population Size.} As shown in Figure~\ref{fig:evolution_population}, we compare Spearman's $\rho$ across initial population sizes of 80, 100, and 200. The data indicate a positive correlation between population size and the initial Spearman: larger initial populations yield higher Spearman's $\rho$ at the outset. 

\noindent\textbf{Number of Unary.} Table~\ref{tab:ablation_unary_number} presents an ablation study examining the effect of unary operation counts on Spearman's rank correlation coefficient and winning rate. The study shows that a lower number of unary operations (2) yields the highest Spearman correlation (86.48\%) and winning rate (25.61\%), indicating that large unary operations may lead to over-complex proxies.

\section{Conclusion}

In this paper, we present LPZero, an innovative approach for discovering proxies for language models without  training or expert intervention. Our LPZero encompasses the design of a comprehensive proxy search space, spanning existing ZC proxies. With genetic programming, we efficiently unearth promising ZC proxies within this space. To expedite the search, we propose a \textit{Rule-based Pruning Strategy}, eliminating less promising proxies early in the process. 
To ascertain the efficacy of LPZero, we conducted experiments on the FlexiBERT and GPT-2 benchmarks to evaluate the ranking consistency of the searched proxy, demonstrating the superior ranking capabilities of LPZero. Furthermore, we assessed LPZero’s performance on commonsense reasoning tasks, where it exhibited commendable results.

\section*{Acknowledgements}

This work was partially supported by National Natural Science Foundation of China under Grant No. 62272122, the Guangzhou Municipal Joint Funding Project with Universities and Enterprises under Grant No. 2024A03J0616, the Hong Kong RIF grant under Grant No. R6021-20, and Hong Kong CRF grants under Grant No. C2004-21G and C7004-22G. 

\section*{Limitations}
This study undertakes a comprehensive review of existing Zero-cost (ZC) proxies specifically tailored for Transformer architectures, integrating them into a unified framework for evaluation. By benchmarking these ZC proxies within the FlexiBERT and GPT-2 benchmarks, we rigorously assess their ranking capabilities through Kendall's $\tau$ and Spearman's $\rho$. This approach allows us to present a systematic comparison of their effectiveness in identifying promising language model architectures without the need for extensive computational resources. Our evaluation focuses on the architectural aspects of language models, aiming to streamline the search process for efficient and effective neural network designs.

However, it's important to note that our research primarily concentrates on the structural design and optimization of language models, sidelining enhancements in specific functional areas such as inference capabilities, logical analysis, advanced language generation, nuanced natural language understanding, and retrieval and integration of knowledge. These critical components of language model performance and applicability in real-world applications are not directly addressed by our current framework. Recognizing these gaps, we identify substantial opportunities for future research to delve into these aspects. Expanding the scope of Zero-cost proxy evaluation to include these functionalities could significantly elevate the utility and comprehensiveness of language models, offering a more holistic approach to their development and assessment in the field of artificial intelligence.

\section*{Ethics Statement}
Our LPZero framework addresses the technical development of language model architectures, sidestepping direct ethical or social considerations. Our work is likely to increase the adoption of NAS in the NLP domain, providing an economic way to perform estimation in language models. 

Despite this focus, we recognize that the application of our findings—aimed at reducing computational demands and streamlining language model development—could intersect with broader ethical issues in natural language processing, such as data privacy, algorithmic bias, and the potential for misuse. We advocate for future research to integrate ethical considerations, scrutinize training data sources for biases, and ensure the responsible deployment of language models, acknowledging their profound societal impact. 
We acknowledge the significant capabilities and prospects offered by artificial intelligence, particularly ChatGPT, in refining written materials. As we utilize this technology to enhance paragraphs, we pledge to adhere strictly to the utmost ethical guidelines, thereby guaranteeing the preservation of integrity, the respect of intellectual property rights, and the support of inclusivity. It is important to clarify that our use of ChatGPT is limited to the refinement of existing content rather than the generation of new content for the paper.


\bibliography{custom}
\bibliographystyle{acl_natbib}

\newpage
\appendix

\section*{Appendix Overview}

\begin{itemize}
    \item Section~\ref{sec:app:proxy_ss}: Details of Proxy Search Space. 
    \item Section~\ref{app:searched_proxy}: Details of the Searched Proxies. 
    \item Section~\ref{app:efficiency_zc_proxies}: Efficiency of Zero-cost Proxies. 
    \item Section~\ref{app:predefined}: Predefined Criteria in RPS. 
    \item Section~\ref{sec:app:correlation_rank}: Rank Correlation. 
    \item Section~\ref{sec:app:unary}: Ablation Study of Unary Operations. 
    \item Section~\ref{app:flexibert_exp}: Additional Experiments on FlexiBERT Benchmark. 
    \item Section~\ref{sec:app:relatedwork}: Additional Related Work. 
    \item Section~\ref{sec:app:lms}: Additional Experiments on more Language Models.
    \item Section~\ref{sec:app:previous_auto}: Comparison of LPZero with Previous Automatic Methods.     
\end{itemize}

\section{Details of Proxy Search Space}\label{sec:app:proxy_ss}

\paragraph{Details of Primitive Operations}

Table~\ref{tab:operation_vocabulary} presents a set of primitive operations used in our framework, which includes 20 unary operations and four binary operations. The unary operations cover a wide range of mathematical functions, such as logarithmic, exponential, trigonometric, and statistical operations, as well as activation functions commonly used in neural networks. The binary operations include basic arithmetic operations: addition, subtraction, multiplication, and division. Each operation is defined by its input and output argument types (scalar, vector, or matrix) and the corresponding mathematical equation. The input and output arguments are denoted using a memory addressing scheme, where the subscript represents the memory address.

\begin{table*}[h]
\centering
\resizebox{\linewidth}{!}{
\begin{tabular}{c|c|c|c|c}
\toprule
\textbf{Op ID} & \textbf{Symbols} & \textbf{Input Args} & \textbf{Output Args} & \textbf{Description} \\
& & \textbf{Addresses/types} & \textbf{Address/type} & \textbf{(in equations)} \\ 
\midrule
$f_{01}$ & log(s1) & $a$ / scalar,vector,matrix & $b$ / scalar,vector,matrix & $y_b = \log(x_a)$ \\
$f_{02}$ & abs(log(s1)) & $a$ / scalar,vector,matrix & $b$ / scalar,vector,matrix & $y_b = |\log(x_a)|$ \\
$f_{03}$ & abs(s1) & $a$ / scalar,vector,matrix & $b$ / scalar,vector,matrix & $y_b = |x_a|$ \\
$f_{04}$ & square(s1) & $a$ / scalar,vector,matrix & $b$ / scalar,vector,matrix & $y_b = (x_a)^{2}$ \\
$f_{05}$ & exp(s1) & $a$ / scalar,vector,matrix & $b$ / scalar,vector,matrix & $y_b = e^{x_a}$ \\
$f_{06}$ & sqrt(s1) & $a$ / scalar,vector,matrix & $b$ / scalar,vector,matrix & $y_b = \sqrt{x_a}$ \\
$f_{07}$ & relu(s1) & $a$ / scalar,vector,matrix & $b$ / scalar,vector,matrix & $y_b = \max(0, x_a)$ \\
$f_{08}$ & reciprocal(s1) & $a$ / scalar,vector,matrix & $b$ / scalar,vector,matrix & $y_b = \frac{1}{x_a}$ \\
$f_{09}$ & neg(s1) & $a$ / scalar,vector,matrix & $b$ / scalar,vector,matrix & $y_b = -{x_a}$ \\
$f_{10}$ & norm\_fro(s1) & $a$ / vector,matrix & $b$ / scalar & $y_b = ||x_a||_F$ \\
$f_{11}$  & norm\_sum(s1) & $a$ / vector,matrix & $b$ / scalar & $y_b = \frac{\sum_i^N{x_a^i}}{\text{numel}(x_a)}$ \\
$f_{12}$ & norm\_l1(s1) & $a$ / vector,matrix & $b$ / scalar & $y_b = ||x_a||_1$ \\
$f_{13}$  & softmax(s1) & $a$ / vector,matrix & $b$ / vector,matrix & $y_b = \frac{e^{x_a^i}}{\sum_{j=1}^{n} e^{x_a^j}}$ \\
$f_{14}$ & sigmoid(s1) & $a$ / vector,matrix & $b$ / vector,matrix & $y_b = \frac{1}{1 + e^{-x_a^i}}$ \\
$f_{15}$ & log\_softmax(s1) & $a$ / vector,matrix & $b$ / vector,matrix & $y_b = \text{log}(f_{12}(x_a))$ \\
$f_{16}$ & min-max scaling(s1) & $a$ / vector,matrix & $b$ / vector,matrix & $y_b = \frac{(x - \min(x_a))}{\max(x_a) - \min(x_a)}$ \\
$f_{17}$ & average(s1) & $a$ / vector,matrix & $b$ / scalar & $y_b = \frac{\sum_i^N{x_a^i}}{N}$ \\
$f_{18}$ & std(s1) & $a$ / vector,matrix & $b$ / scalar & $y_b = \sqrt{\frac{1}{N}\sum_{i=1}^{N}(x_a^i - \mu)^2}$ \\
$f_{19}$ & s1 & $a$ / scalar,vector,matrix & $b$ / scalar,vector,matrix & $y_b = x_a$ \\
$f_{20}$ & $\varnothing$ & - & - &$y_b = \varnothing$ \\
\midrule
$g_{01}$ & add(s1,s2) & $a,b$ / scalars,vectors,matrices & $c$ / scalars,vectors,matrices & $y_c = x_a + x_b$ \\
$g_{02}$ & sub(s1,s2) & $a,b$ / scalars,vectors,matrices & $c$ / scalars,vectors,matrices & $y_c = x_a - x_b$ \\
$g_{03}$ & mul(s1,s2) & $a,b$ / scalars,vectors,matrices & $c$ / scalars,vectors,matrices & $y_c = x_a \cdot x_b$ \\
$g_{04}$ & div(s1,s2) & $a,b$ / scalars,vectors,matrices & $c$ / scalars,vectors,matrices & $y_c = x_a / x_b$ \\
\bottomrule
\end{tabular}
}
\caption{Primitive operation set $\mathcal{K}$. Summary of unary (denoted by $f$) and binary Operations (denoted by $g$).}
\label{tab:operation_vocabulary}
\end{table*}

\paragraph{Details of FlexiBERT}
\label{sec:app:flexibert}

Table~\ref{app:tab:detail_flexibert} provides a detailed overview of the FlexiBERT benchmark, highlighting the diverse range of hyperparameters available for tuning. FlexiBERT, designed to explore architectural variations within the BERT model framework, allows for configurations that span the specifications of BERT-Tiny and BERT-Mini. Key architectural elements are outlined along with their corresponding hyperparameter values, including hidden dimension sizes, the number of encoder layers, types of attention operators, and more. Notably, the hidden dimension and the number of encoder layers are consistent across the architecture, whereas other parameters vary across encoder layers, introducing a high degree of flexibility and customization. The table also specifies the conditions under which different attention operation parameters are applied, depending on the type of attention operator selected. With a total of 10,621,440 possible architectures, this proxy search space represents a comprehensive framework for exploring and identifying efficient model configurations within the BERT architecture spectrum.

\begin{table*}[t]
\centering
\small 
\begin{tabular}{lc}
\toprule
\textbf{Architecture Element} & \textbf{Hyperparameters Values} \\
\midrule
Hidden dimension & \{128, 256\} \\ 
Number of Encoder Layers & \{2, 4\} \\ 
Type of attention operator & \{self-attention, linear, span-based dynamic convolution\} \\ 
Number of operation heads & \{2, 4\} \\ 
Feed-forward dimension & \{512, 1024\} \\ 
Number of feed-forward stacks & \{1, 3\} \\ 
Attention operation parameters & \begin{tabular}[c]{@{}c@{}}if self-attention \{scaled dot-product, multiplicative\} \\ if linear transform \{discrete Fourier, discrete cosine\} \\ if dynamic convolution convolution kernel size: \{5, 9\}\end{tabular} \\
\bottomrule
\end{tabular}
\caption{The FlexiBERT benchmark, with hyperparameter values spanning those found in BERT-Tiny and BERT-Mini. Hidden dimension and number of encoder layers is fixed across the whole architecture; all other parameters are heterogeneous across encoder layers. The benchmark encompasses 10,621,440 architectures.}\label{app:tab:detail_flexibert}
\end{table*}

\paragraph{Details of GPT-2}\label{sec:app:gpt-2}

Table~\ref{app:tab:detail_gpt2} delineates the expansive benchmark leveraged for the GPT-2 architecture optimization, outlining a comprehensive set of hyperparameters targeted in the exploration process. It includes the number of layers (nlayer), representing the depth of the transformer model; the dimensionality of model embeddings (dmodel), indicative of the scale and capacity of the model; the inner dimension of the feed-forward networks (dinner), a critical parameter for the model's ability to process and integrate information within each transformer layer; the number of attention heads (nhead), which impacts the model's ability to focus on different parts of the input sequence; and the dimensions of adaptive input embeddings (dembed) along with their associated scaling factors (k), parameters that offer a novel approach to managing input representation complexity and efficiency. A noteworthy aspect of this benchmark is the adaptive setting of the dinner parameter, which is dynamically adjusted to be at least twice the dmodel size, a heuristic introduced to mitigate the risk of training collapse by ensuring sufficient capacity in the feed-forward networks.

\begin{table*}[t]
\centering
\small 
\begin{tabular}{lc}
\toprule 
\textbf{Architecture Element} & \textbf{Hyperparameters Values} \\
\midrule 
Number of Layers (nlayer) & $\{2, 3, ..., 16\}$ \\
Model Dimension (dmodel) & $\{128, 192, ..., 1024\}$ \\
Inner Dimension (dinner) & $\{256, 320, ..., 4096\}$ \\
Number of Attention Heads (nhead) & $\{2, 4, 8\}$ \\
Adaptive Input Embedding Dimension (dembed) & $\{128, 256, 512\}$ \\
Adaptive Input Embedding Factor (k) & $\{1, 2, 4\}$ \\
\bottomrule
\end{tabular}
\caption{The GPT-2 benchmark, covering a broad spectrum of architectural configurations. Once a model dimension (dmodel) is chosen, the minimum inner dimension (dinner) is set to twice the value of dmodel to avoid training collapse. This adaptive approach ensures a wide range of effective and efficient architectures, summing up to more than $10^{54}$ unique configurations.}\label{app:tab:detail_gpt2}
\end{table*}

\paragraph{Details of LLaMA}\label{sec:app:llama_ss}

\begin{figure*}[h]
    \centering
    \includegraphics[width=0.7\linewidth]{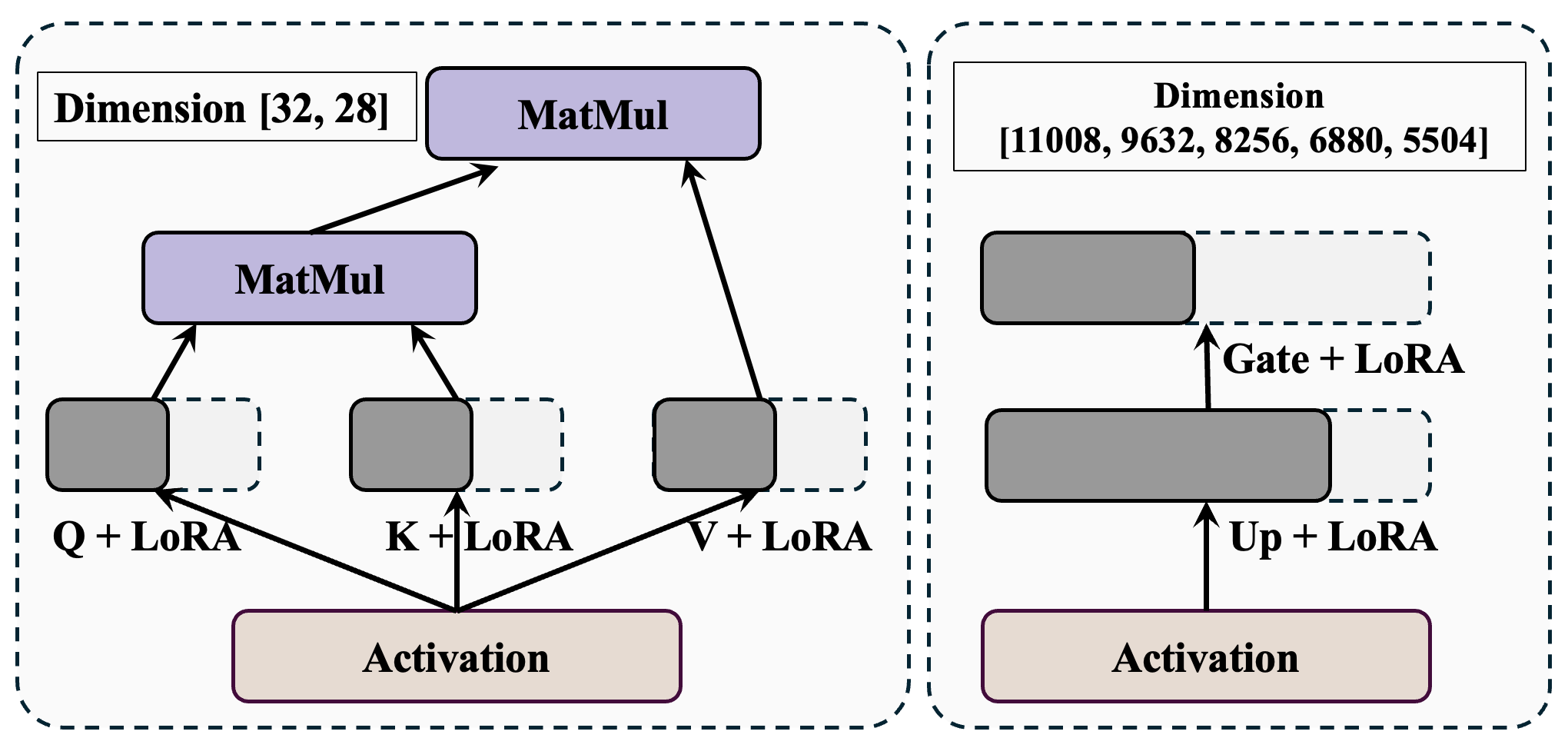}
    \caption{Illustration of the search space for the LLaMA super-network. The left diagram depicts the multi-head attention mechanism, where the query (Q), key (K), and value (V) matrices are augmented with LoRA modules of dimension [32, 28]. The right diagram represents the feed-forward network (FFN) component, where the up and gate matrices are enhanced with LoRA modules of dimension [11008, 9632, 8256, 6880, 5504].}
    \label{fig:app:llama}
\end{figure*}

In this paper, we employ the same search space on LLaMA as LoNAS~\cite{munoz-etal-2024-lonas-elastic}. To convert the pre-trained LLMs like LLaMA into supernet, LoNAS proposed elastic low-rank adapters to explore the search space. The details of the search space are presented in Figure~\ref{fig:app:llama}. It presents a comprehensive illustration of the search space for the LLaMA super-network, which consists of two primary components: the multi-head attention mechanism and the feed-forward network (FFN).

The left diagram showcases the multi-head attention mechanism, where the input is first transformed into query (Q), key (K), and value (V) matrices through linear transformations. These matrices are then augmented with low-rank adaptation (LoRA) modules, denoted as Q + LoRA, K + LoRA, and V + LoRA, respectively. The dimensions of these LoRA modules are fixed at [32, 28], indicating a search space that allows for the exploration of different rank values within this range. The augmented matrices undergo the standard attention computation, which involves matrix multiplication and softmax operations, to produce the output of the multi-head attention mechanism.

The right diagram focuses on the FFN component of the LLaMA super-network. In this part, the input undergoes a series of transformations through the up and gate matrices. These matrices are enhanced with LoRA modules, represented as Up + LoRA and Gate + LoRA, respectively. The search space for these LoRA modules is defined by the dimensions [11008, 9632, 8256, 6880, 5504], allowing for the exploration of various rank configurations within this range. The output of the Up + LoRA matrix passes through an activation function, while the Gate + LoRA matrix acts as a gating mechanism to control the flow of information. The outputs from both branches are then combined to form the final output of the FFN component.

In LLaMA-7B based supernet, we treat all 31 transformer layers equally. The search space size is determined by the possible combinations of LoRA module dimensions for each layer. With 2 configurations per layer for the multi-head attention mechanism and 5 configurations per layer for the FFN component, the total search space size is $2^{31} \times 5^{31}$. This large search space allows for a comprehensive exploration of different LoRA module settings to find the optimal configuration that maximizes performance while minimizing additional parameters.

\section{Details of the Searched Proxies}\label{app:searched_proxy}

\paragraph{LPZero Proxy for FlexiBERT Benchmark} The mathematical formulation of the searched ZC proxy on FlexiBERT Benchmark is given by:
\begin{equation*}
    \varphi(\theta_{H}, \theta_{A}) = \sum_{i=0}^{N} ((\frac{1}{\theta_{H}})^2 + \log\left(\eta\left(||\theta_{A}||_F\right)\right))
\end{equation*}
where \(\theta_{H}\) denotes the parameters associated with the heads in the Multi-head Attention, \(\theta_{A}\) represents the activation values of each block within the network, and \(\eta\) symbolizes the softmax operation.

The formulated Zero-cost (ZC) proxy equation effectively evaluates neural architectures by considering both their structural efficiency and functional performance. The first term prioritizes models with fewer, yet efficient, parameters in the attention mechanism \(\left(\frac{1}{\theta_{H}}\right)^2\), highlighting the goal of Zero-shot NAS towards computational efficiency. The second term \(\log\left(\eta\left(\|\theta_{A}\|_F\right)\right)\) focuses on the diversity and distribution of activations, aiming for architectures that ensure balanced and effective information processing. Together, these aspects form a comprehensive approach for the holistic evaluation of architectures in the FlexiBERT benchmark, which is critical for identifying optimal models for NLP tasks.

\paragraph{LPZero Proxy for GPT-2 Benchmark} The mathematical formulation of the searched ZC proxy on GPT-2 Benchmark is given by:
\begin{align*}
    \varphi(\theta_G, \theta_W) &= \sum_{i=1}^{N} \left(|\text{normalize}(\theta_G)|\right. \\    
    &\quad + \left. \log\left(\left|\text{mean}(\theta_W)|\right)\right|\right) 
\end{align*}
where \(\theta_G\) denotes the parameters associated with the generator within the GPT-2 architecture, and \(\theta_W\) represents the weights of each layer within the network.

The formulated Zero-cost (ZC) proxy equation effectively evaluates neural architectures by considering both their structural efficiency and functional performance. The first term \(|\text{normalize}(\theta_G)|\) emphasizes the significance of normalized generator parameters, highlighting the importance of parameter scaling and stability in the generation process, which is critical for computational efficiency. The second term \(\log\left(\left|\text{mean}(\theta_W)\right|\right)\) focuses on the average weight magnitudes, aiming for architectures that ensure balanced weight distributions and effective learning dynamics. Together, these aspects form a comprehensive approach for the holistic evaluation of architectures in the GPT-2 benchmark, which is essential for identifying optimal models for language generation tasks.

\paragraph{LPZero Proxy for LLaMA Benchmark} The mathematical formulation of the searched ZC proxy on LLaMA Benchmark is given by:
\begin{equation*}
\begin{split}
    \varphi(\theta_{W_1}, \theta_{W_2}) = \sum_{i=1}^{N} \Big(&\|\theta_{W_1}\|_1^{2} \\
    &+ \left(\text{softmax}\,\theta_{W_2}\right)^{\frac{1}{2}}\Big)
\end{split}
\end{equation*}
where \(\theta_{W_1}\) denotes the first set of weight parameters in the LLaMA architecture, and \(\theta_{W_2}\) represents the second set of weight parameters.

The formulated Zero-cost (ZC) proxy equation effectively evaluates neural architectures by considering both their structural efficiency and functional performance. The first term \(\left(\|\theta_{W_1}\|_1^{2}\right)\) emphasizes the importance of the \(L_1\)-norm squared of the first set of weight parameters, which encourages sparsity and leads to more efficient models. The second term \(\left(\text{softmax}(\theta_{W_2})\right)^{\frac{1}{2}}\) focuses on the softmax-transformed second set of weights, aiming to ensure balanced weight distributions and effective scaling. Together, these aspects form a comprehensive approach for the holistic evaluation of architectures in the LLaMA benchmark, which is critical for identifying optimal models for language modeling tasks.

\section{Efficiency of Zero-cost Proxies}\label{app:efficiency_zc_proxies}

We evaluate the efficiency of various zero-cost proxies by measuring their average evaluation time on the FlexiBERT benchmark. The results are presented in Table~\ref{tab:app:efficiency}, which compares the evaluation time and Kendall Tau correlation of each proxy. Among the proxies tested, Synaptic Diversity exhibits the fastest evaluation time at 0.672 seconds, followed closely by Activation Distance at 0.754 seconds. However, both of these proxies demonstrate relatively low Kendall Tau correlations of 0.021 and 0.081, respectively, indicating a weaker relationship between their rankings and the actual performance of the architectures. On the other hand, our proposed method, LPZero, achieves the highest Kendall Tau correlation of 0.511, suggesting a strong agreement between its rankings and the true performance rankings. This superior correlation comes at the cost of a longer evaluation time of 3.818 seconds, which is still competitive with other high-performing proxies such as LogSynflow and Synflow. Our proposed LPZero method serves as an efficient alternative to evaluating the performance of architectures on downstream tasks, which can be highly time-consuming. By leveraging LPZero as a zero-cost proxy, we can effectively rank and compare different architectures without the need for extensive evaluation on specific tasks.

\begin{table}[t]
    \centering
    \resizebox{\linewidth}{!}{%
    \begin{tabular}{lcc}
    \toprule 
    \textbf{Model} & \textbf{Evaluation Time} & \textbf{Kendall Tau} \\
    \midrule 
    Synaptic Diversity & 0.672s & 0.021 \\ 
    Activation Distance & 0.754s & 0.081 \\ 
    Head Importance & 0.908s & 0.050 \\ 
    Jacobian Cosine & 0.916s & 0.116 \\ 
    Attention Confidence & 0.920s & 0.475 \\ 
    GradNorm & 0.936s & 0.133 \\ 
    Synaptic Saliency & 0.944s & 0.157 \\ 
    SNIP & 0.976s & 0.119 \\ 
    GraSP & 1.626s & 0.122 \\ 
    Fisher & 1.794s & 0.139 \\ 
    \gr LPZero (Ours) & 3.818s & \textbf{0.511} \\ 
    LogSynflow & 5.306s & 0.334 \\ 
    Synflow & 5.376s & 0.322 \\
    \bottomrule 
    \end{tabular}}
    \caption{Comparison of evaluation time on FlexiBERT Benchmark of different ZC proxies.}
    \label{tab:app:efficiency}
\end{table}

\section{Predefined Criteria in RPS}\label{app:predefined}

In mathematics, understanding the relationships between various operations significantly impacts the LPZero search space. Table~\ref{tab:app:rps} summarizes the relationships among a set of operations, categorizing them based on their mathematical interactions. These relationships include inverse functions, derivatives, equivalence, special cases, and potential conflicts when certain operations are combined. This overview helps in recognizing how operations can complement or conflict with each other, thereby providing support for RPS.

\begin{table*}[t]
\centering
\resizebox{\textwidth}{!}{
\begin{tabular}{cccl}
\toprule 
\textbf{OP 1} & \textbf{OP 2} & \textbf{Relationship} & \textbf{Description} \\ \midrule 
log & exp & Inverse & $e^x$ and log(x) are inverse functions. \\
abs & abs(log) & Derivative & Absolute value operation applied to log. \\
square & sqrt & Inverse & Squaring and square root are inverse operations. \\
ReLU & identity & Special case & ReLU acts as identity for x>0. \\
inverse & identity & Inverse for non-zero & Multiplicative inverse operation. \\
norm\_sum & average & Equivalent & Norm sum divided by count is average. \\
softmax & log\_softmax & Derivative & Log softmax is the logarithm of softmax. \\
sigmoid & logistic function & Equivalent & Sigmoid is also known as the logistic function. \\
min-max scaling & normalization & Type & Min-max scaling is a type of normalization. \\
standard deviation & variance & Square root & Standard deviation is the square root of variance. \\
L1-norm & abs & Generalization & L1-norm is a sum of absolute values. \\
F-norm & Euclidean norm & Equivalent & Frobenius norm for matrices, Euclidean norm for vectors. \\
-() & log & Conflict & Negation followed by log leads to undefined result for positive inputs. \\
-() & sqrt & Conflict & Negation followed by sqrt leads to undefined result for positive inputs. \\
identity (if zero) & inverse & Conflict & Identity ensuring zero input followed by inverse leads to division by zero. \\
-() (for positive) & sqrt & Conflict & Negation of positive numbers followed by sqrt is undefined. \\ \bottomrule
\end{tabular}}
\caption{Summary of Predefined Criteria}\label{tab:app:rps}
\label{tab:my_label}
\end{table*}


\section{Rank Correlation}\label{sec:app:correlation_rank}

As a complement to the visualization of ranking correlation, we follow LiteTransformerSearch~\cite{Javaheripi2022LiteTransformerSearchTN} and provide visualizations of GLUE Score Ranking and ZC Proxies Ranking. It can be observed that potential proxies are capable of dividing the candidate models into two clusters at least through ranking. This further demonstrates the robustness of our LPZero results.

\begin{figure*}[t]
    \centering
    \includegraphics[width=1\linewidth]{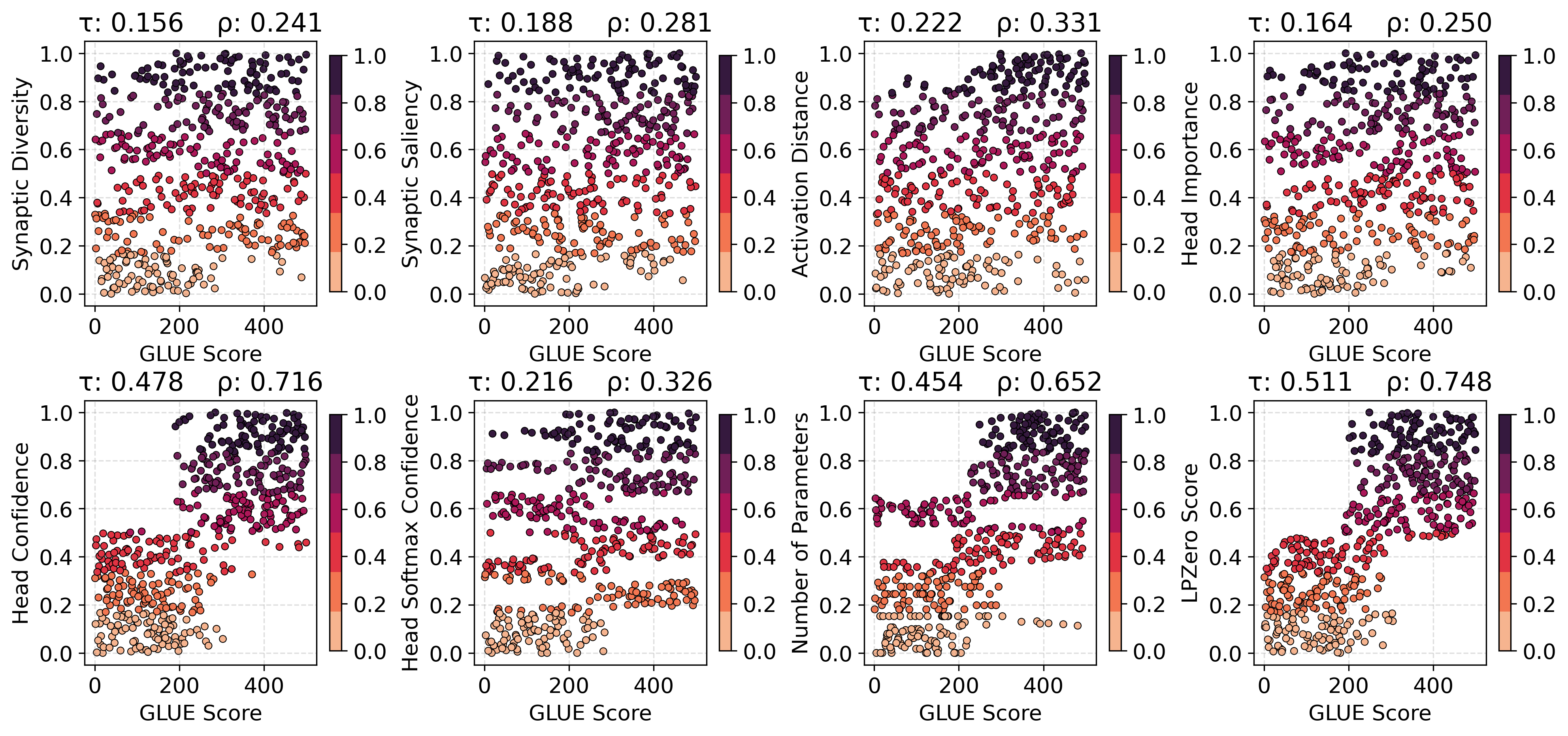}
    \caption{Correlation of training-free proxies ranking with GLUE Ranking on 500 architectures randomly sampled from FlexiBERT benchmark. }
    \label{fig:correlation_ranking}
\end{figure*}

\section{Ablation Study of Unary Operations}\label{sec:app:unary}

\begin{figure*}[t]
    \centering
    \begin{minipage}[b]{0.49\linewidth}
        \includegraphics[width=\linewidth]{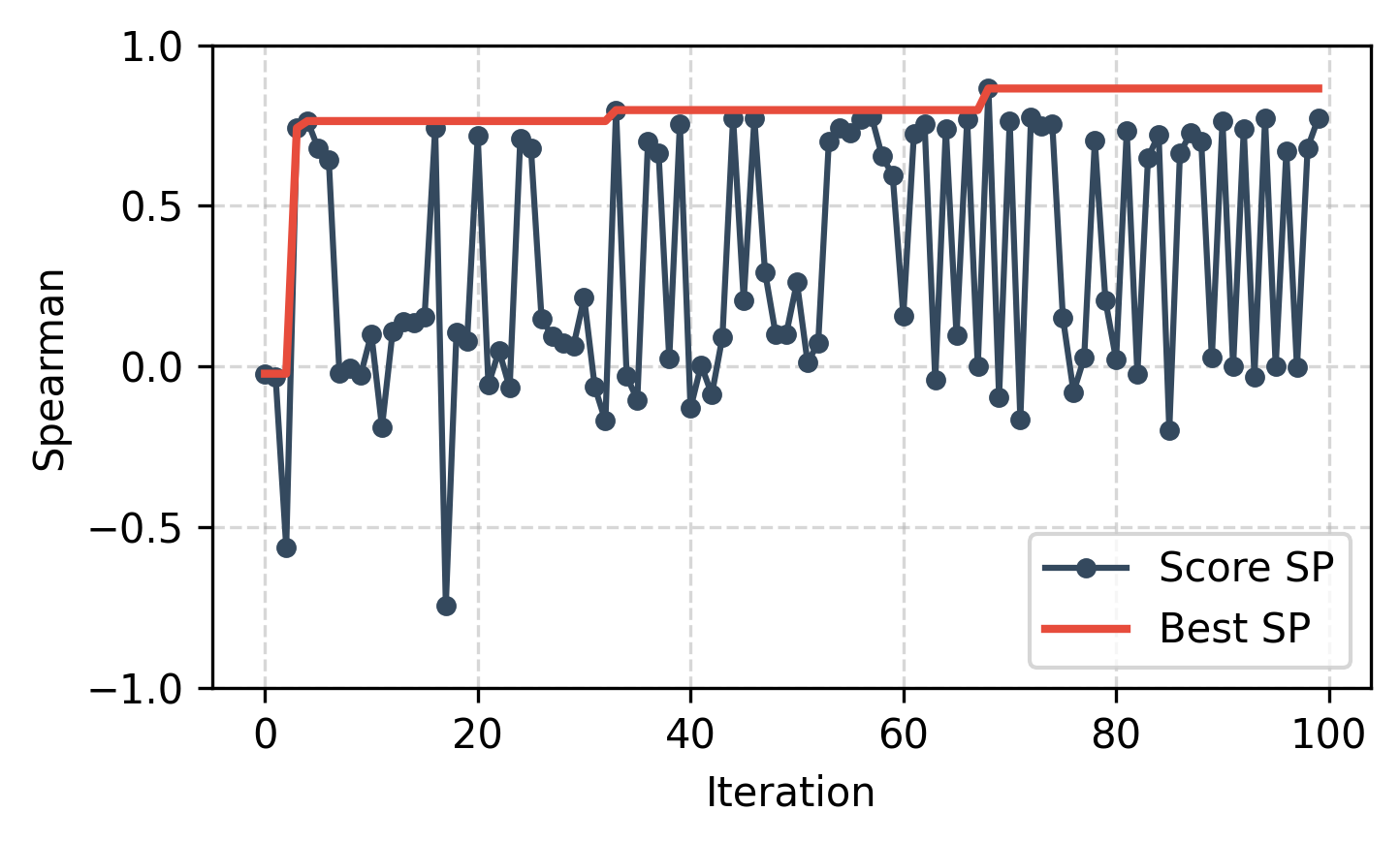}
        \caption{Two Unary Operations.}
        \label{fig:label1}
    \end{minipage}
    \hfill 
    \begin{minipage}[b]{0.49\linewidth}
        \includegraphics[width=\linewidth]{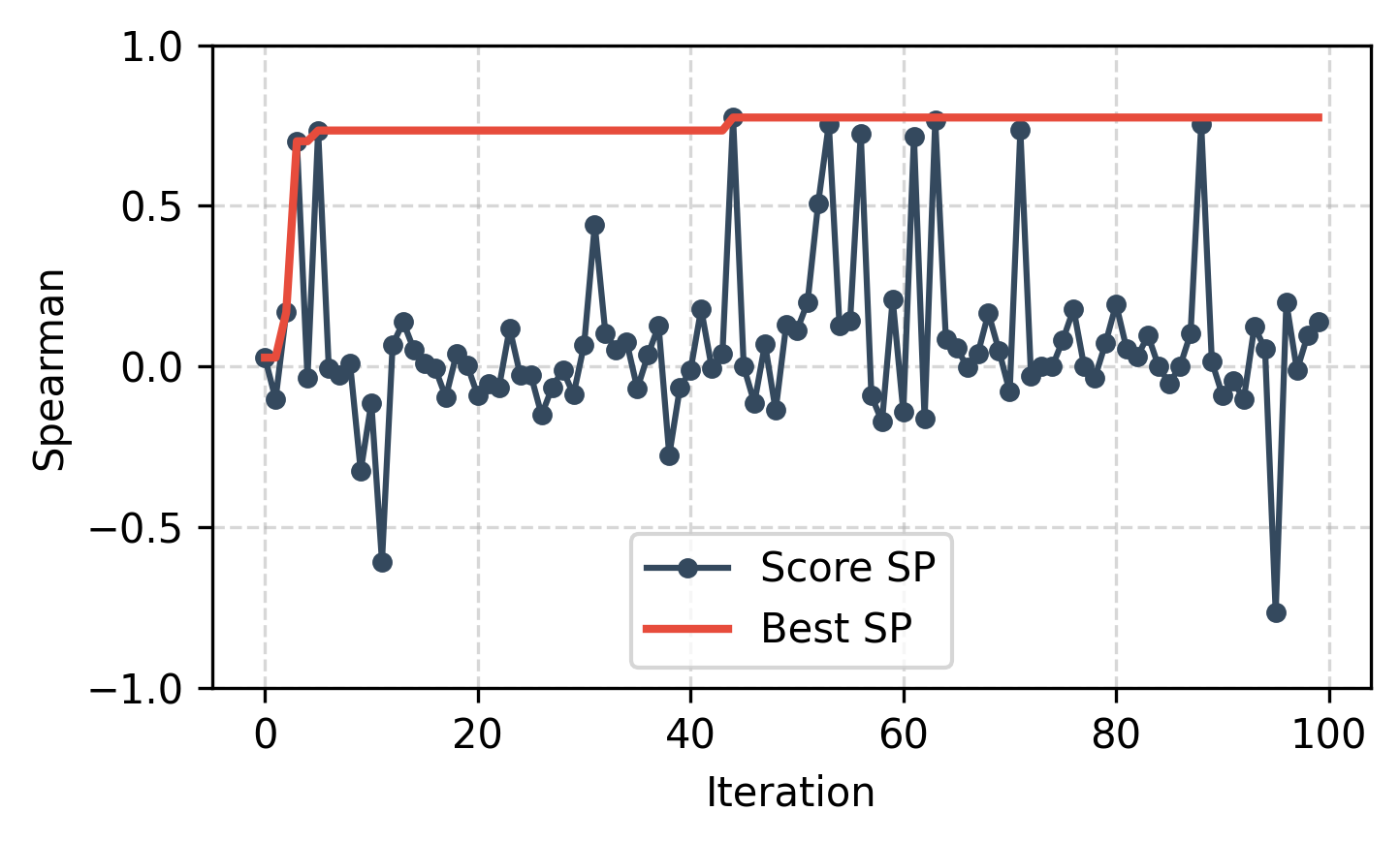}
        \caption{Three Unary Operations.}
        \label{fig:label2}
    \end{minipage}

    \begin{minipage}[b]{0.49\linewidth}
        \includegraphics[width=\linewidth]{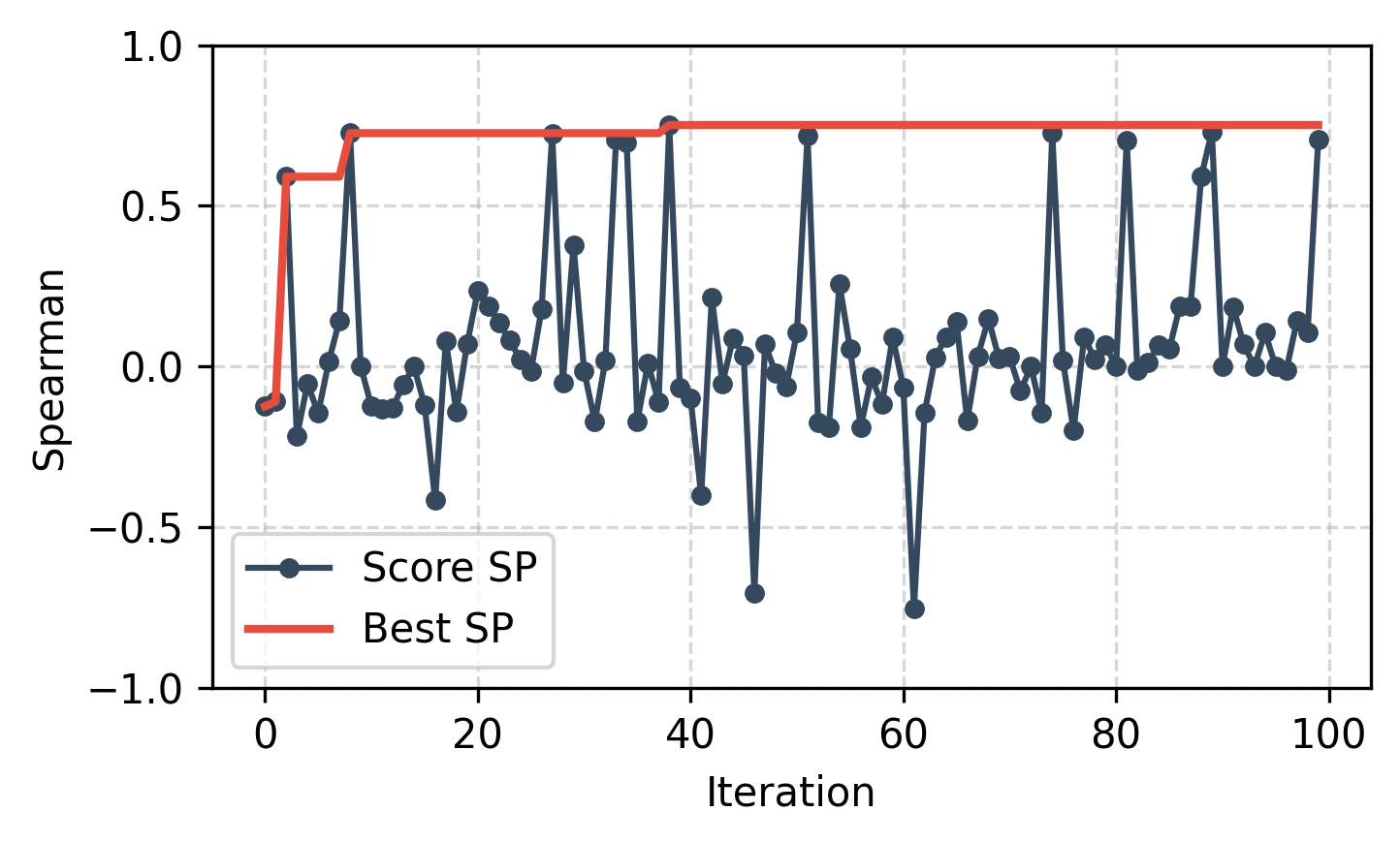}
        \caption{Four Unary Operations.}
        \label{fig:label3}
    \end{minipage}
    \hfill 
    \begin{minipage}[b]{0.49\linewidth}
        \includegraphics[width=\linewidth]{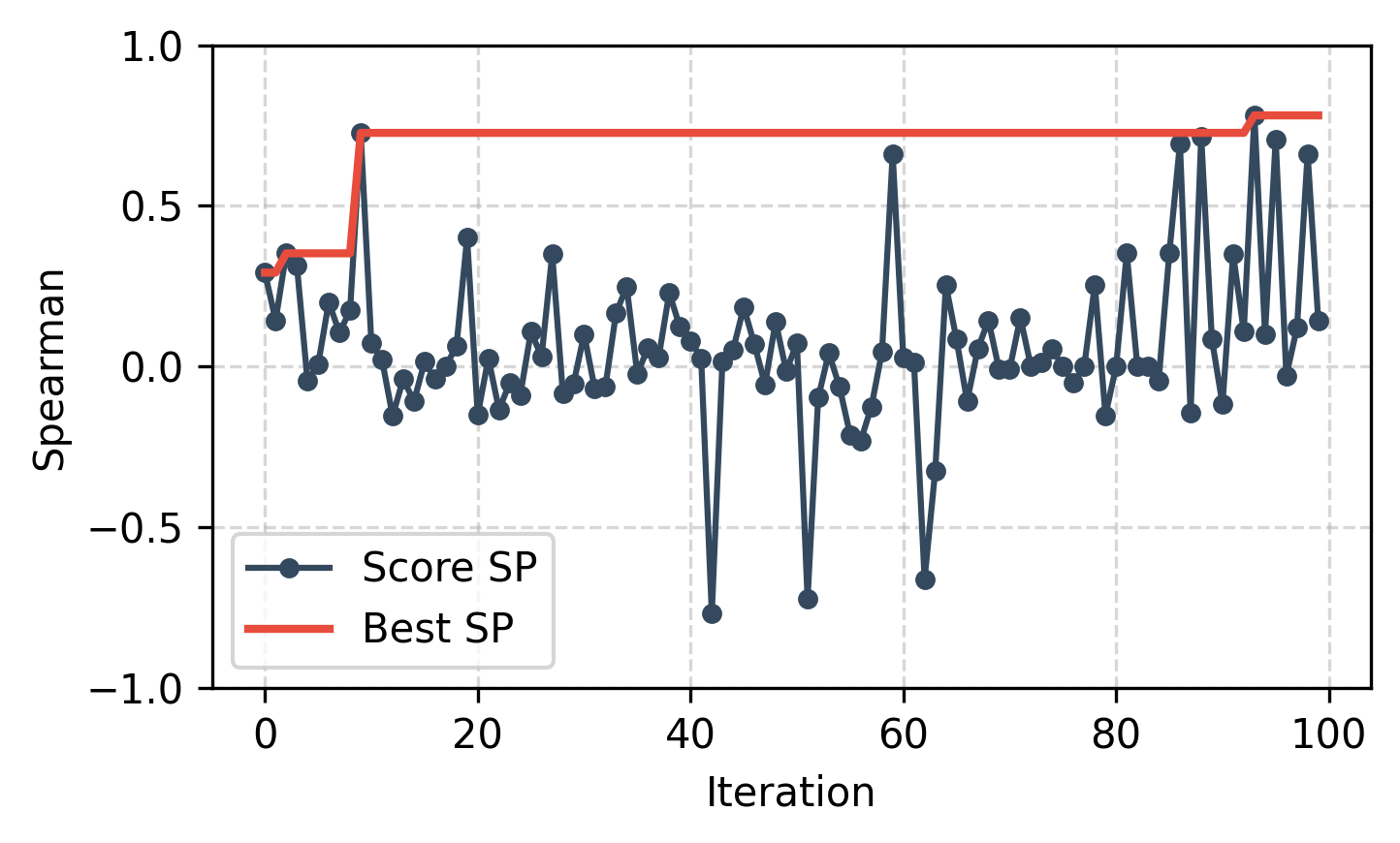}
        \caption{Five Unary Operations.}
        \label{fig:label4}
    \end{minipage}
    
    \caption{Ablation Study of the Number of Unary Operations.}
    \label{fig:overall-label}
\end{figure*}

Figure~\ref{fig:overall-label} illustrates an ablation study that investigates the performance of systems with varying unary operations. It presents four graphs, each plotting performance metrics in 100 iterations for systems with two to five unary operations. The study finds that the system with two unary operations achieves and maintains the highest 'Best SP' score, indicating stable, optimal performance. Systems with more than two unary operations show more fluctuations in 'Best SP' and a lower Spearman rank correlation, suggesting that additional operations may lead to over-complexity and reduced performance. Thus, the optimal number of unary operations for this system is two, balancing complexity and performance.

\section{Additional Experiments on FlexiBERT Benchmark}\label{app:flexibert_exp}

As presented in Table~\ref{tab:app:flexibert_additional_exp}, the LPZero model demonstrated superior performance with the highest average score of 76.57 among the tested zero-cost (ZC) proxies. Notably, LPZero excelled particularly in the SST-2 task, achieving a top score of 85.32, underscoring its effectiveness in sentiment analysis. In contrast, models such as GraSP and Activation Distance lagged significantly, with average scores of 64.40 and 65.51 respectively, indicating challenges in tasks requiring sophisticated linguistic understanding. The performance disparity across models highlights the importance of proxy selection based on task-specific characteristics, suggesting that while LPZero offers robust general performance, other proxies may require further refinement to enhance their effectiveness across diverse NLP tasks.

\begin{table*}[t]
    \centering
    \resizebox{\textwidth}{!}{%
    \begin{tabular}{lcccccccc}
    \toprule 
    \textbf{Model} & \textbf{QNLI} & \textbf{MRPC} & \textbf{SST-2} & \textbf{CoLA} & \textbf{STS-B} & \textbf{MNLI-m} &  \textbf{QQP} & \textbf{AVG} \\
    \midrule 
    \#Params. & 82.28 & 80.39 & 82.68 & 43.24 & 83.99 & 72.94 & 86.55 & 76.01 \\
    Fisher~\cite{Turner2019BlockSwapFB_fisher} & 80.8 & 70.34 & 82.22 & 23.08 & 83.85 & 66.91 & 83.86 & 70.15 \\
    GradNorm~\cite{abdelfattah2021zerocost} & 81.99 & 79.9 & 82.11 & 27.55 & 84.68 & 70.95 & 85.59 & 73.25 \\
    GraSP~\cite{Wang2020PickingWT_GraSP} & 78.77 & 72.79 & 78.90 & 0.0 & 81.28 & 63.93 & 82.13 & 64.40 \\
    LogSynflow~\cite{cavagnero2022freerea} & 82.96 & 81.62 & 83.72 & 42.69 & 83.72 & 72.57 & 86.90 & 76.31 \\
    Synflow~\cite{tanaka2020pruning_synflow} & 81.22 & 75.00 & 79.93 & 5.79 & 83.8 & 67.23 & 83.80 & 68.11 \\
    SNIP~\cite{Lee2018SNIPSN} & 82.28 & 80.39 & 82.68 & 43.24 & 83.99 & 72.94  & 86.55 & 76.01 \\
    Synaptic Saliency~\cite{tanaka2020pruning_synflow} & 82.81 & 82.84 & 83.83 & 33.45 & 84.49 & 70.40 & 85.28 & 74.43 \\
    Jacobian Cosine~\cite{celotti2020improving} & 78.11 & 71.57 & 76.95 & 9.35 & 79.11 & 62.28 & 81.32 & 65.53 \\
    Attention Confidence~\cite{serianni-kalita-2023-training} & 81.00 & 76.47 & 81.08 & 30.21 & 83.67 & 68.36 & 84.69 & 72.21 \\
    Activation Distance~\cite{mellor2021neural} & 79.52 & 73.04 & 76.95 & 0.0 & 82.67 & 64.25 & 82.16 & 65.51 \\
    Synaptic Diversity~\cite{DSS} & 82.96 & 81.62 & 83.72 & 42.69 & 84.59 & 72.57 & 86.90 & 76.43 \\
    Head Importance~\cite{serianni-kalita-2023-training} & 81.33 & 79.90 & 80.05 & 30.90 & 84.23 & 67.44 & 84.24 & 72.58 \\
    \gr LPZero (Ours) & 83.34 & 82.35 & 85.32 & 40.80 & 84.08 & 73.45 & 86.68 & \textbf{76.57} \\
    \bottomrule 
    \end{tabular}}
    \caption{Comparison of results on FlexiBERT Benchmark of different ZC proxies.}
    \label{tab:app:flexibert_additional_exp}
\end{table*}

\section{Additional Related Work}\label{sec:app:relatedwork}

\noindent\textbf{Activation Distance} Activation Distance, specifically in the context of NWOT~\cite{mellor2021neural}, leverages binary activation patterns to measure the correlation between input data across ReLU (Rectified Linear Unit) layers within a neural network. This proxy is crucial for understanding how different inputs activate the network's architecture, providing insights into the diversity and richness of the learned representations. The formula provided,
\begin{equation}
    \mathcal{S}=\log|K_H|
\end{equation}
where \(K_H\) represents the kernel matrix, quantifies the similarity (or distance) between activation patterns. The determinant of the kernel matrix (\(|K_H|\)) captures the volume of the space spanned by the activations, and taking its logarithm transforms this volume measure into a more manageable scale. 

\noindent\textbf{Synaptic Saliency}
Synaptic Saliency, or Synflow~\cite{tanaka2020pruning_synflow}, is a criterion used to identify the importance of parameters (weights) in a neural network, aiming to approximate the impact on the loss function when a specific parameter is removed. This concept is framed within the equation,
\begin{equation}
    \mathcal{S}=\frac{\partial \mathcal{L}}{\partial \mathcal{\theta}} \odot \theta
\end{equation}
where \(\frac{\partial \mathcal{L}}{\partial \mathcal{\theta}}\) denotes the gradient of the loss function with respect to the parameters (\(\mathcal{\theta}\)), and \(\odot\) represents the Hadamard product, signifying element-wise multiplication between the gradient and the parameters themselves. This approach to quantifying parameter importance is designed to prevent layer collapse during the pruning process of network training, ensuring that the pruning does not disproportionately affect any single layer which could result in significant performance degradation.

\noindent\textbf{Jacobian Score Cosine}
The Jacobian Score Cosine (JSC)~\cite{celotti2020improving} is a Zero-cost Proxy designed to evaluate the sensitivity and stability of neural network architectures with respect to their input data. By analyzing the Jacobian matrix, which represents the first derivatives of the network's outputs with respect to its inputs, the JSC offers insights into how small variations in the input can affect the output, thereby assessing the network's robustness and generalization capability. The JSC is computed using the following formula:
\begin{equation}
    S = 1 - \frac{1}{N^2-N} \sum_{i=1}^{N} \left[ J_n J_n^t - I \right]^{\frac{1}{20}},
\end{equation}
where \(S\) denotes the Jacobian Score, \(N\) is the number of inputs to the network, \(J_n\) represents the Jacobian matrix for the \(n\)th input, \(J_n^t\) is the transpose of \(J_n\), and \(I\) is the identity matrix. This equation calculates the average cosine similarity between the Jacobian vectors of all pairs of inputs, adjusted by the identity matrix to normalize self-similarity, and finally raised to the power of \(\frac{1}{20}\) to scale the measure.

\noindent\textbf{Synaptic Diversity} The concept of \textit{Synaptic Diversity} within the context of Training-Free Transformer Architecture Search (TF-TAS)~\cite{DSS} represents a novel approach towards evaluating and selecting Vision Transformer (ViT) architectures. By circumventing the need for extensive training, this methodology significantly enhances computational efficiency in Transformer Architecture Search (TAS). The TF-TAS scheme, delineated in the studies by Zhou et al., employs a modular strategy that assesses ViT architectures through two theoretical lenses: synaptic diversity and synaptic saliency, collectively referred to as the DSS-indicator.

Synaptic Diversity, particularly in relation to multi-head self-attention (MSA) modules of ViTs, is instrumental in gauging the performance of these architectures. This proxy evaluates the heterogeneity of synaptic connections by utilizing the Nuclear-norm as an approximate measure for the rank of weight matrices within MSA modules. A higher Nuclear-norm indicates a greater diversity, which suggests a potential for enhanced performance due to the ability to encapsulate a broader spectrum of features and relationships within the data. The computation of Synaptic Diversity is formalized as follows:
\begin{equation}
    S = \sum_m \left\| \frac{\partial \mathcal{L}}{\partial W_m} \right\| \odot \|W_m\|_{\text{nuc}}
\end{equation}
Here, \(S\) symbolizes the synaptic diversity score, \(\frac{\partial \mathcal{L}}{\partial W_m}\) denotes the gradient of the loss function with respect to the weights of the \(m\)-th MSA module, and \(\|W_m\|_{\text{nuc}}\) is the Nuclear-norm of the weight matrix, serving as a proxy for the rank and thus the diversity of the synaptic connections.

\noindent\textbf{Hidden Covariance} The Hidden Covariance proxy provides a sophisticated means to analyze the behavior and interaction of hidden states within a specific layer of a Recurrent Neural Network (RNN) when processing a minibatch of \(N\) input sequences \(X = \{x_n\}_{n=1}^N\). This proxy is particularly insightful for examining the internal dynamics and dependencies of the hidden states across different time steps or sequences. Given the hidden state matrix \(H(X)\) for a minibatch, we first compute the covariance matrix \(C\) as follows:
\begin{equation}
C = (H - M_H)(H - M_H)^T,
\end{equation}
\noindent where \(M_H\) is the mean matrix derived from the hidden states, with its elements defined by:
\begin{equation}
(M_H)_{ij} = \frac{1}{N} \sum_{n=1}^N H_{in},
\end{equation}
\noindent indicating the average activation across the minibatch for each hidden unit. This step captures the variance and covariance of the hidden states, highlighting the variability and correlation of activations in response to the input batch. Subsequently, to normalize and interpret the covariance values, we calculate the Pearson product-moment correlation coefficients matrix \(R\) as:
\begin{equation}
R_{ij} = \frac{C_{ij}}{\sqrt{C_{ii}C_{jj}}},
\end{equation}
\noindent which standardizes the covariance matrix into a correlation matrix \(R\), providing a normalized measure of linear dependencies between pairs of hidden units.

Building upon the framework established by \citet{mellor2021neural}, the final proxy \(S(H)\) is derived using the Kullback–Leibler divergence from the eigenvalues of the kernel of \(R\), computed as:
\begin{equation}
S(H) = - \sum_{n=1}^N \left( \log(\lambda_n + k) + \frac{1}{\lambda_n + k} \right),
\end{equation}
where \(\lambda_1, \dots, \lambda_N\) are the eigenvalues of \(R\), and \(k = 10^{-5}\) is a small constant added to stabilize the logarithm and reciprocal operations.

Note that Hidden Covariance is designed for RNN architectures, which means it is not working for Transformer-based networks. That is why we don not report the performance of Hidden Covariance on FlexiBERT and GPT-2 benchmark. 

\noindent\textbf{Confidence} The Confidence proxy~\cite{serianni-kalita-2023-training} quantifies the average maximum attention (or activation) that a neural network layer, specifically an attention mechanism, directs towards the most significant features or tokens for a set of inputs $X$. This is mathematically articulated as:
\begin{equation}
\mathcal{S} = \frac{1}{N} \sum_{n=1}^N \max(\text{Att}(h, x_n))
\end{equation}
In this expression, $\mathcal{S}$ symbolizes the average maximal attention score across all instances within the minibatch, where $\text{Att}(h, x_n)$ signifies the attention scores calculated for the $n$-th input by the function $h$. 

\begin{table*}[t]
\centering
\small 
\caption{Comparison of LPZero with its counterparts.}
\label{tab:ml_optimization}
\begin{tabularx}{\textwidth}{l|XXXXX}
\toprule 
\textbf{Description} & \textbf{AutoML-Zero~\shortercite{real2020automlzero}} & \textbf{EZNAS~\shortercite{akhauri2022eznas}} & \textbf{Auto-Prox~\shortercite{wei2024auto}} & \textbf{EMQ~\shortercite{dong2023emq}} & \textbf{LPZero} \\ \midrule
\textbf{Task} & Machine Learning Program Discovery & Zero-shot NAS & Zero-shot NAS & Mixed-precision Quantization & Symbolic Expression \\ \hline
\textbf{Targets} & - & CNN & ViT & CNN & LLMs \\ \hline
\textbf{Params} & - & 0.3-1.5MB & 2-25MB & 13.4-44.6MB & 7B \\ \hline
\textbf{Retrain} & No & Yes & Yes & Yes & No \\ \hline
\textbf{Strategy} & Evolutionary Algorithm & Distributed Evolutionary Algorithm in Python (DEAP) & Elitism-Preserve Strategy & Diversity Prompting Selection & Rule-based Pruning Strategy \\ \hline
\textbf{Objective} & Find machine learning algorithms from scratch & Find optimal proxy that can measure the convolution-based architectures & Find the optimal proxy that can measure the vit-based architectures & Find the optimal metric that can better rank candidate bit-width configurations & Find the optimal symbolic equation that can predict the performance of LLMs \\ 
\bottomrule
\end{tabularx}
\end{table*}

\noindent\textbf{Softmax Confidence} Softmax Confidence~\cite{serianni-kalita-2023-training} broadens the notion of Confidence to scenarios where softmax scores, derived from the softmax function $\sigma$, are utilized to gauge the network's prediction certainty. The formulation is given by:
\begin{equation}
\mathcal{S} = \frac{1}{N} \sum_{n=1}^N \max(\sigma(h, x_n))
\end{equation}
Here, $\sigma(h, x_n)$ computes the softmax probabilities for the outputs related to the $n$-th input, and the $\max$ operation selects the highest probability, denoting the model's most confident prediction for each input. The mean of these maxima across the minibatch offers a measure of the overall prediction confidence, valuable for assessing the certainty of classification decisions by the model.

\noindent\textbf{Importance} The \textbf{Importance} proxy \cite{serianni-kalita-2023-training} assesses the sensitivity of the cost function $\mathcal{C}(X)$ with respect to the attention mechanism $\text{Att}_h(X)$ for a given input set $X$. This sensitivity analysis is crucial for understanding the impact of changes in attention weights on the overall performance or cost of the neural network. The Importance proxy is mathematically represented as:
\begin{equation}
\mathcal{S} = \left| \frac{\partial \mathcal{C}(X)}{\partial \text{Att}_h(X)} \right|
\end{equation}
This equation calculates the absolute value of the derivative of the cost function relative to the attention weights, quantifying the "importance" of the attention mechanism in the network's decision-making process. A higher value suggests that minor adjustments to the attention weights could lead to significant changes in the cost, underscoring the critical areas of the input that the network focuses on.

\noindent\textbf{SNIP} (Single-shot Network Pruning)~\cite{Lee2018SNIPSN} introduces a pruning criterion that can be applied early in the training process, even before the actual training commences. It is predicated on the sensitivity of the loss function $\mathcal{L}$ with respect to each parameter $\theta$, modulated by the parameter values themselves. The SNIP criterion is formulated as:
\begin{equation}
    S(\theta) = \left| \frac{\partial \mathcal{L}}{\partial \theta} \odot \theta \right|
\end{equation}
where the operation $\odot$ denotes the element-wise product. This expression evaluates the absolute value of the gradient of the loss function with respect to the parameters, weighted by the parameters themselves. This criterion aids in identifying parameters that have minimal impact on the loss function, allowing for their pruning to streamline the model architecture without significantly compromising performance.

\noindent\textbf{GraSP} \textbf{(Gradient Signal Preservation)}~\cite{Wang2020PickingWT_GraSP} introduces a pruning methodology aimed at preserving the gradient flow throughout the network's architecture. This strategy identifies and eliminates parameters that have the least effect on the gradient flow, thus minimizing their impact on the network's ability to learn. The GraSP criterion is quantitatively defined by the equation:
\begin{equation}
    S(\theta) = - \left( H \frac{\partial \mathcal{L}}{\partial \theta} \right) \odot \theta
\end{equation}
In this formulation, \(S(\theta)\) denotes the pruning score assigned to each parameter \(\theta\), reflecting its significance in maintaining effective gradient flow within the network. The term \(H\) represents the Hessian matrix, which consists of the second-order derivatives of the loss function \(\mathcal{L}\) with respect to the parameters, while \(\frac{\partial \mathcal{L}}{\partial \theta}\) is the gradient of the loss with respect to the parameters. The operation \(\odot\) signifies element-wise multiplication, and the negative sign indicates that parameters which contribute negatively to the gradient flow—and therefore potentially hinder learning—are prioritized for removal.

The principal insight of GraSP is its emphasis on the Hessian-gradient product, which offers a measure of the influence of parameter changes on the curvature of the loss landscape and, subsequently, on the dynamics of model training. By focusing on preserving parameters critical for the integrity of gradient flow, GraSP enables network pruning in a manner that is less likely to degrade performance.

In this paper, we have chosen not to incorporate the Hessian Matrix as part of our analysis due to its computationally intensive nature. However, it is worth noting that excluding considerations of computational load, the inclusion of the Hessian Matrix could potentially enhance performance significantly.

\begin{table*}[t]
\centering
\resizebox{\textwidth}{!}{
\begin{tabular}{lcccccccccc}
\toprule
\textbf{Method} & \textbf{Params.} & \textbf{Cost} & \textbf{BoolQ} & \textbf{PIQA} & \textbf{HellaSwag} & \textbf{WinoGrande} & \textbf{Arc-e} & \textbf{Arc-c} & \textbf{OBQA} & \textbf{Avg.} \\
\midrule
LLaMA-1 & 6.7B & - & 73.18 & 78.35 & 72.99 & 67.01 & 67.45 & 41.38 & 42.40 & 63.25 \\
SliceGPT~\cite{slicegpt_iclr24} & 5.0B & \textbf{\textasciitilde 5} & - & 66.87 & 63.38 & 54.16 & 58.46 & 34.56 & - & 55.49 \\
LLM-Pruner~\cite{Ma2023LLMPrunerOT} & 5.4B & \textbf{3} & 69.54 & 76.44 & 68.11 & 65.11 & 63.43 & 37.88 & 40.00 & 59.79 \\
FLAP~\cite{an2023fluctuationbased} & 5.5B & \textbf{\textasciitilde 1} & 69.63 & 76.82 & 71.20 & 68.35 & 69.91 & 39.25 & 39.40 & 62.08 \\
Shortened LLaMA~\cite{kim2024shortened} & 5.5B & \textbf{2,688} & 72.70 & 75.70 & 70.40 & 63.60 & 69.50 & 40.19 & 41.20 & 61.89 \\
LLaMA-NAS~\cite{sarah2024llama} & 5.0B & \textbf{-} & - & - & - & 63.00 & 67.2 & 40.60 & - & 56.93 \\
SLEB~\cite{Song2024SLEBSL} & 5.5B & \textbf{-} & - & 73.07 & 58.96 & 62.47 & 56.48 & 33.02 & - & 56.80 \\
LPZero (Ours) & 5.7B & \textbf{-} & 62.17 & 74.81 & 52.65 & 65.43 & 74.62 & 57.42 & 62.17 & 64.18 \\
\bottomrule
\end{tabular}
}
\caption{A comparative analysis of LLaMA-1 based pruning methodologies, including their respective parameters, additional post-training computational costs (measured in GPU hours), and performance metrics across various benchmarks.}
\label{tab:comparison-llama1}
\end{table*}

\begin{table*}[t]
\centering
\resizebox{.95\textwidth}{!}{
\begin{tabular}{lcccccccc}
\toprule
\textbf{} & \textbf{Params.} & \textbf{Cost} & \textbf{PIQA} & \textbf{WinoGrande} & \textbf{Arc-c} & \textbf{Arc-e} & \textbf{Avg.} \\
\midrule
LLaMA-2 & 7B & - & 78.80 & 69.06 & 41.89 & 74.58 & 66.08 \\
LLM-Pruner~\cite{Ma2023LLMPrunerOT}(30\%) & 4.9B & \textbf{3} & 71.81 & 54.06 & 30.30 & 63.42 & 54.89 \\
SliceGPT~\cite{slicegpt_iclr24}(20\%) & 5.6B & \textbf{1} & 69.42 & 65.59 & 37.54 & 71.84 & 61.09 \\
OSP~\cite{Gao2024OptimizationbasedSP}(30\%) & 4.9B & \textbf{2.7} & 75.41 & 61.60 & 35.58 & 66.31 & 61.10 \\
LLaMA-NAS~\cite{sarah2024llama} & 5B & \textbf{-} & - & 63.00 & 40.60 & 67.2 & 56.93 \\
Wanda 2:4 ~\cite{Sun2023ASA_wanda}& - & \textbf{-} & 70.84 & 62.27 & 31.97 & 57.58 & 55.65 \\
SLEB~\cite{Song2024SLEBSL} & 5.5B & \textbf{-} & 73.07 & 58.96 & 33.02 & 56.48 & 55.38 \\
LPZero (Ours) 5.7B & 5.7B & \textbf{-} & 72.69 & 57.14 & \textbf{57.34} & \textbf{75.04} & \textbf{65.55} \\
\bottomrule
\end{tabular}
}
\caption{A comparative analysis of LLaMA-2 based pruning methodologies, detailing their respective parameters, the additional post-training computational costs (measured in GPU hours), and their performance metrics across various benchmarks.}
\label{tab:comparison-llama2}
\end{table*}

\noindent\textbf{LogSynflow} ~\cite{cavagnero2022freerea} introduces a nuanced variation to the conventional pruning criteria by applying a logarithmic transformation to the gradients' magnitude. This adjustment is intended to enhance the pruning strategy by ensuring a more nuanced evaluation of parameter importance, especially for those with small but significant gradients. The LogSynflow criterion is mathematically expressed as:
\begin{equation}
    S(\theta) = \theta \cdot \left| \log \left| \frac{\partial \mathcal{L}}{\partial \theta} \right| \right|
\end{equation}
In this equation, \(S(\theta)\) represents the score assigned to each parameter \(\theta\) based on its importance, where \(\frac{\partial \mathcal{L}}{\partial \theta}\) denotes the gradient of the loss function \(\mathcal{L}\) with respect to the parameters. The use of the absolute value of the logarithm of the gradient magnitude aims to highlight the significance of parameters that might otherwise be overlooked due to their relatively small gradient values. By multiplying these logarithmic values by the parameters themselves, LogSynflow prioritizes the retention of parameters that are integral to the network's ability to learn, thereby facilitating a more informed pruning process that minimizes the loss of critical information.

\section{Additional Experiments on more Language Models}\label{sec:app:lms}

We add more details over whether the method has been retrained, particularly regarding the additional post-training cost used in these methods. To clarify these disparities, we present Table~\ref{tab:comparison-llama1} and ~\ref{tab:comparison-llama2} to include descriptions of the additional post-training costs.
Several of the baseline methods (SliceGPT~\cite{slicegpt_iclr24}, LLM-Pruner~\cite{Ma2023LLMPrunerOT}, FLAP~\cite{an2023fluctuationbased}, and Shortened LLaMA~\cite{kim2024shortened}) include additional post-training steps that enhance their performance. These structured pruning methods have incorporated additional post-training, giving them a potential advantage in performance. Below are the details:

\begin{itemize}
    \item \textbf{SliceGPT~\cite{slicegpt_iclr24}}: Uses the standard PEFT method LoRA to recover the performance of fine-tuning (termed RFT in SliceGPT) after structured pruning. Specifically, SliceGPT employed 8,000 samples from Alpaca for fine-tuning, which costs around five GPU hours.
    \item \textbf{LLM-Pruner~\cite{Ma2023LLMPrunerOT}}: Provided LoRA post-trained results. Even for LLM-Pruner, our method achieves better performance than it.
    \item \textbf{FLAP~\cite{an2023fluctuationbased}}: Did not incorporate LoRA for post-training after pruning but proposed a new method called Baseline Bias Compensation, serving a function similar to LoRA fine-tuning. From the paper of FLAP, we find that they only provide the data utilized in the pruning process, which is 1,024 samples with around one GPU hour.
    \item \textbf{Shortened LLaMA~\cite{kim2024shortened}}: Incorporated a tedious post-training process after pruning, with 2,688 GPU hours to get the final model.
\end{itemize}

Similar to LoNAS~\cite{munoz-etal-2024-lonas-elastic}, our method has no additional post-training process using LoRA. Similar to LoNAS, LoRA is utilized for creating the supernet. Our results are more competitive when considering only those methods that do not involve additional post-training costs. We will clarify the differences of these methods in the revision. We highlight (in bold) the highest score for each task when additional post-training cost was not involved.

Additionally, we present the results on LLaMA-2 on Table~\ref{tab:comparison-llama2}. Specifically, we further investigated the recent published papers (including OSP~\cite{Gao2024OptimizationbasedSP}, LLaMA-NAS~\cite{sarah2024llama}, Wanda 2:4~\cite{Sun2023ASA_wanda}) about structured pruning in the following table. We list their performance as well as their additional post-training cost. Our LPZero can achieve competitive performance for LLaMA-2 as base model. We highlight (in bold) the highest score for each task when additional post-training cost was not involved.

\section{Comparison of LPZero with Previous Automatic Methods}\label{sec:app:previous_auto}

We compare our proposed method, LPZero, with previous automatic methods for proxy searching, including AutoML-Zero~\cite{real2020automlzero}, EZNAS~\cite{akhauri2022eznas}, Auto-Prox~\cite{wei2024auto}, and EMQ~\cite{dong2023emq}. The comparison is presented in Table~\ref{tab:ml_optimization}, which highlights the key differences and improvements of LPZero over its counterparts.

The table compares various aspects of these methods, such as the task they address, the target models they optimize, the number of parameters in the target models, whether retraining is required, the optimization strategy employed, and the objective of each method. LPZero stands out from the other methods in several ways. First, it focuses on optimizing large language models (LLMs) with up to 7 billion parameters, which is significantly larger than the target models of other methods. Second, LPZero does not require retraining the model, which can be computationally expensive and time-consuming. Instead, it employs a novel rule based pruning strategy to find the optimal symbolic equation that can predict the performance of LLMs. In contrast, AutoML-Zero aims to discover machine learning algorithms from scratch, while EZNAS~\cite{akhauri2022eznas} and Auto-Prox~\cite{wei2024auto} focus on neural architecture search (NAS) for convolutional neural networks (CNNs) and vision transformers (ViTs), respectively. EMQ~\cite{dong2023emq} tackles the problem of mixed-precision quantization~\cite{Lin2024DuQuantDO} for CNNs. By providing this comparative analysis, we highlight the unique contributions and advantages of LPZero in the context of automatic machine learning optimization methods, particularly its ability to handle large-scale models and its efficient optimization strategy that eliminates the need for retraining. 
Future work could explore the integration of LPZero with a variety of AutoML techniques~\cite{dong2023emq,Dong2023diswot,Pruner-Zero,wei2024auto,li2024attnzero,li2024Autogas} to enhance model selection and hyperparameter tuning. Additionally, combining LPZero with distillation methods~\cite{lishadow,li2024kd,li2022self,li2022norm,li2024detkds} could lead to more efficient model compression while maintaining accuracy.
Incorporating quantization techniques~\cite{Frantar2022GPTQAP,Xiao2022SmoothQuantAA,du-etal-2024-bitdistiller} may further optimize model inference by reducing size and computational demands.

\end{document}